\documentclass[numbers]{article}

    \usepackage[final]{neurips_2024}

\usepackage[utf8]{inputenc} 
\usepackage[T1]{fontenc}    
\usepackage{hyperref}       
\usepackage{url}            
\usepackage{booktabs}       
\usepackage{amsfonts}       
\usepackage{nicefrac}       
\usepackage{microtype}      
\usepackage{xcolor}         
\newtheorem{theorem}{Theorem}[section]

\newtheorem{lemma}[theorem]{Lemma}

\newtheorem{definition}[theorem]{Definition}
\newtheorem{assumption}[theorem]{Assumption}

\usepackage{algorithm}
\usepackage{algorithmic}
\renewcommand{\algorithmicrequire}{\textbf{Input:}} 
\renewcommand{\algorithmicensure}{\textbf{Output:}} 

\usepackage{subfig}
\usepackage{colortbl}
\usepackage{multirow}
\usepackage{amsmath}
\usepackage{graphicx}
\usepackage{amssymb}
\usepackage{wrapfig}
\usepackage{multirow}
\usepackage{multicol}
\hypersetup{
colorlinks = true,
linkcolor = black,
citecolor = black
}

\title{Heterogeneity-Guided Client Sampling: Towards Fast and Efficient Non-IID Federated Learning}

\author{%
  Huancheng Chen\\
  University of Texas at Austin\\
  \texttt{huanchengch@utexas.edu} \\
  \And
  Haris Vikalo \\
  University of Texas at Austin\\
  \texttt{hvikalo@ece.utexas.edu}
}

\begin{document}

\maketitle

\begin{abstract}
  Statistical heterogeneity of data present at client devices in a federated learning (FL) system renders the training of a global model in such systems difficult. Particularly challenging are the settings where due to communication resource constraints only a small fraction of clients can participate in any given round of FL. Recent approaches to training a global model in FL systems with non-IID data have focused on developing client selection methods that aim to sample clients with more informative updates of the model. However, existing client selection techniques either introduce significant computation overhead or perform well only in the scenarios where clients have data with similar heterogeneity profiles. In this paper, we propose HiCS-FL (Federated Learning via Hierarchical Clustered Sampling), a novel client selection method in which the server estimates statistical heterogeneity of a client's data using the client’s update of the network’s output layer and relies on this information to cluster and sample the clients. We analyze the ability of the proposed techniques to compare heterogeneity of different datasets, and characterize convergence of the training process that deploys the introduced client selection method. Extensive experimental results demonstrate that in non-IID settings HiCS-FL achieves faster convergence than state-of-the-art FL client selection schemes. Notably, HiCS-FL drastically reduces computation cost compared to existing selection schemes and is adaptable to different heterogeneity scenarios. 
\end{abstract}

\vspace{- 0.1 in}
\section{Introduction}
\vspace{- 0.1 in}
\label{intro}
The federated learning (FL) framework enables privacy-preserving collaborative training of machine learning (ML) models across a number of devices (clients) by avoiding the need to collect private data stored at those devices. The participating clients typically experience both the system as well as statistical heterogeneity \citep{li2020federated}. The former describes settings where client devices have varying degree of computational resources, communication bandwidth and fault tolerance, while the latter refers to the fact that the data owned by the clients may be drawn from different distributions. In this paper, we focus on FL under statistical heterogeneity and leave studies of system heterogeneity to future work. 

An early FL method, FedAvg \citep{fedavg}, performs well in the settings where the devices train on independent and identically distributed (IID) data. However, compared to the IID scenario, training on non-IID data is detrimental to the convergence speed, variance and accuracy of the learned model. This has motivated numerous studies aiming to reduce the variance and improve convergence of FL on non-IID data \citep{fedhkd, collins2021exploiting, scaffold, li2021ditto, fedprox,fedma}. 

On another note, constraints on communication resources and therefore on the number of clients that may participate in training additionally complicate implementation of FL schemes. It would be particularly unrealistic to require regular contributions to training from all the clients in a large-scale cross-device FL system. Instead, only a fraction of clients participate in any given training round; unfortunately, this further aggravates detrimental effects of statistical heterogeneity. 
Selecting informative clients in non-IID FL settings is an open problem that has received considerable attention from the research community \citep{powerofchoice, clustered, afl}. Since privacy concerns typically prohibit clients from sharing their local data label distributions, existing studies focus on estimating informativeness of a client's update by analyzing the update itself. This motivated a family of methods that rely on the norms of local updates to assign probabilities of sampling the clients \citep{chen2020optimal, ribero2020communication}. Aiming to enable efficient use of the available communication and computation resources, another set of methods groups clients with similar data distributions into clusters based on the similarity between clients' model updates \citep{diverse, clustered}. 
Across the board, the existing methods still struggle to deliver desired performance in an efficient manner and cannot distinguish clients with balanced data from the clients with imbalanced data.

In this paper, we consider training a neural network model for \textbf{classification tasks} via federated learning and propose a novel adaptive clustering-based sampling method for identifying and selecting informative clients. The method, referred to as Federated Learning via Hierarchical Clustered Sampling (HiCS-FL), relies on the updates of the (fully connected) output layer in the network to determine how diverse is the clients' data and, based on that, decide which clients to sample. In particular, HiCS-FL enables heterogeneity-guided client selection by utilizing general properties of the gradients of the output layer to distinguish between clients with balanced from those with imbalanced data. Unlike the Clustered Sampling strategies \citep{clustered} where the clusters of clients are sampled uniformly, HiCS-FL allocates different probabilities (importance) to the clusters according to their average estimated data heterogeneity. Numerous experiments conducted on vision datasets FMNIST, CIFAR10, Mini-ImageNet and a NLP dataset \href{https://github.com/649453932/Chinese-Text-Classification-Pytorch}{THUC news} demonstrate that HiCS-FL achieves significantly faster training convergence and lower variance than the competing methods. Finally, we conduct convergence analysis of HiCS-FL and discuss implications of the results.  

In summary, the contributions of the paper include: (1) Analytical characterization of the correlation between local updates of the output layer and the FL clients' data label distribution, along with an efficient method for estimating data heterogeneity; (2) a novel clustering-based algorithm for heterogeneity-guided client selection; (3) extensive simulation results demonstrating HiCS-FL provides significant improvement in terms of convergence speed and variance over competing approaches; and (4) theoretical analysis of the proposed schemes.

\vspace{- 0.1 in}
\section{Background and Related Work}
\vspace{- 0.1 in}
\label{related_work}
Assume the cross-device federated learning setting with $N$ clients, where client $k$ owns private local dataset $\mathcal{B}_{k}$ with $\left|\mathcal{B}_{k}\right|$ samples. The plain vanilla FL considers the objective
\begin{equation}
\label{objective}
    \min_{\mathbf{\theta}} F(\theta) \triangleq \sum_{k = 1}^{N} p_{k}F_{k}(\mathbf{\theta}),
\end{equation}
where $\mathbf{\theta}$ denotes parameters of the global model, $F_{k}(\mathbf{\theta})$ is the loss (empirical risk) of model $\mathbf{\theta}$ on $\mathcal{B}_{k}$, and $p_{k}$ denotes the weight assigned to client $k$, $\sum_{k = 1}^{N} p_{k} = 1$. 
In FedAvg, the weights are set to $p_{k} = \left|\mathcal{B}_{k}\right| / \sum_{i = 1}^{N} \left|\mathcal{B}_{i}\right|$. In training round $t$, the server collects clients' model updates $\mathbf{\theta}_{k}^{t}$ formed by training on local data and aggregates them to update global model as $\mathbf{\theta}^{t+1} = \sum_{k=1}^{N} p_{k} \mathbf{\theta}_{k}^{t}$. 



When an FL system operates under resource constraints, typically only $K \ll N$ clients are selected to participate in any given round of training; denote the set of clients selected in round $t$ by $\mathcal{S}^{t}$. In departure from FedAvg, FedProx \citep{fedprox} proposes an alternative strategy for sampling clients based on a multinomial distribution where the probability of selecting a client is proportional to the size of its local dataset; the global model is then formed as the average of the collected local models $\mathbf{\theta}^{t+1} = \frac{1}{K} \sum_{k\in \mathcal{S}^{t}} \mathbf{\theta}_{k}^{t}$. This sampling strategy is \emph{unbiased} since the the updated global model is on expectation equal to the one obtained by the framework with full client participation as Eq.\ref{objective}. 

AFL \citep{afl} is the first study to utilize local validation loss as a \emph{value} function for computing client sampling probabilities; Power-of-Choice \citep{powerofchoice} takes a step further to propose a greedy approach to sampling clients with the largest local loss. Both of these methods require all clients to compute the local validation loss, which is often unrealistic. To address this problem, FedCor \citep{fedcor} models the local loss by a Gaussian Process (GP), estimates the GP parameters from experiments, and uses the GP model to predict clients' local losses without requiring them to perform validation. In \citep{chen2020optimal}, Optimal Client Sampling scheme aiming to minimize the variance of local updates by assigning sampling probabilities proportional to the Euclidean norm of the updates is proposed. The study in \citep{ribero2020communication} models the progression of model's weights by an Ornstein-Uhlenbeck process and proposes a strategy, optimal under that assumption, for selecting clients with significant weight updates.

The clustering-based sampling method proposed in \citep{clustered} uses cosine similarity \citep{sattler2020clustered} to group together clients with similar local updates, and proceeds to sample one client per cluster in attempt to avoid redundant gradient information. DivFL \citep{diverse} follows the same principle of identifying representative clients but does so by constructing a submodular set and greedily selecting diverse clients. Both of these techniques are computationally expensive due to the high dimension of the gradients that they need to process. 

In general, the overviewed methods either: (1) select diverse clients to reduce redundant information; or (2) select clients with a perceived significant contributions to the global model (high loss, large update or low class-imbalance). Efficient and effective client selection in FL remains an open challenge, motivating the heterogeneity-guided adaptive client selection method presented next.

\vspace{- 0.1 in}
\section{HiCS-FL: Federated Learning via Hierarchical Clustered Sampling}
\vspace{- 0.1 in}
\label{HiCS}
\setlength\intextsep{0pt}
\begin{wrapfigure}[9]{r}{5.5cm}
\centering
\includegraphics[width=0.35\textwidth]{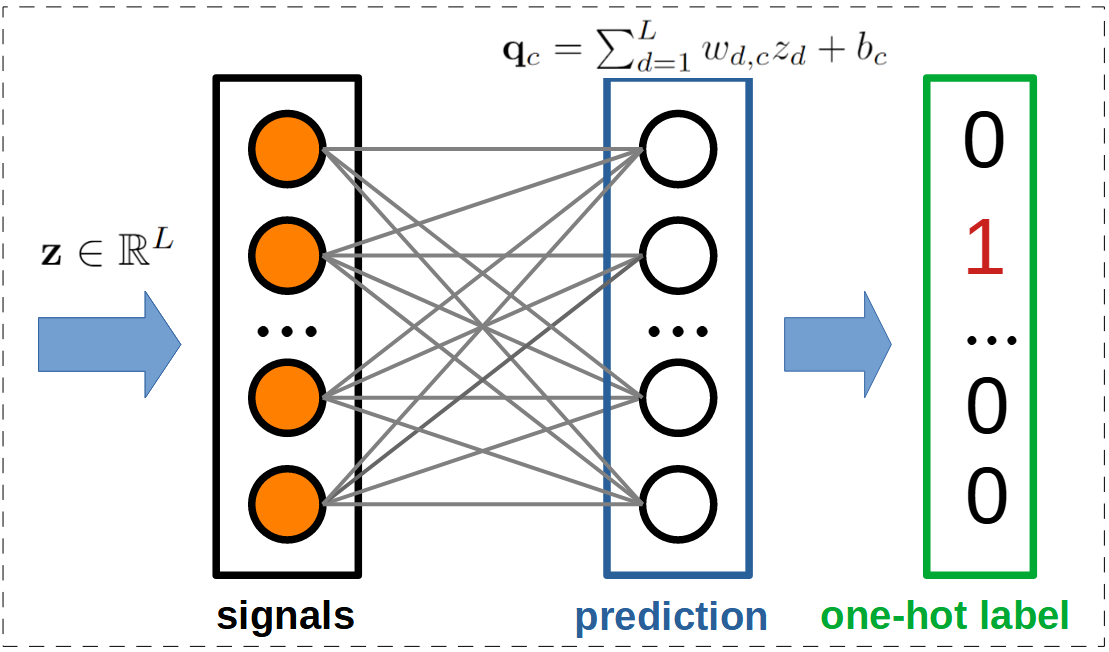}
\caption{The last two network layers.}
\label{fc layer}
\end{wrapfigure}

Existing client sampling methods including Clustered Sampling \citep{clustered} and DivFL \citep{diverse} aim to select clients such that the resulting model update is an unbiased estimate of the true update (i.e., the update in the case of full client participation) while minimizing the variance
\begin{equation}
    \left \Vert \frac{1}{N}\sum_{k = 1}^{N}\nabla F_{k}(\mathbf{\theta}^{t}) - \frac{1}{K}\sum_{k \in \mathcal{S}^{t}}\nabla F_{k} (\mathbf{\theta}^{t}) \right \Vert_{2}^{2}.
\end{equation}
Clustered Sampling, for instance, groups $N$ clients into $K$ clusters based on \emph{representative gradients} \citep{sattler2020clustered}, and randomly selects one client from each cluster to contribute to the global model update. Such an approach unfortunately fails to differentiate between model updates formed on data with balanced and those formed on data with imbalanced label distributions -- indeed, in either case the updates are treated as being equally important. However, a number of studies in centralized learning has shown that class-imbalanced datasets have significant detrimental effect on the performance of learning classification tasks \citep{buda2018systematic,chawla2002smote,shen2016relay}. 
This intuition carries over to the FL settings where one expects the updates from clients training on relatively more balanced local data to have a more beneficial impact on the performance of the system. The Federated Learning via Hierarchical Clustered Sampling (HiCS-FL) framework described in this section adapts to the clients' data heterogeneity in the following way: if the levels of heterogeneity (as quantified by the entropy of data label distribution) vary from one cluster to another, HiCS-FL is more likely to sample clusters containing clients with more balanced data; if the clients grouped in different clusters have similar heterogeneity levels, HiCS-FL is more likely to select diverse clients (i.e., sample uniformly across clusters, thus reducing to the conventional clustered sampling strategy).

\vspace{- 0.1 in}
\subsection{Class-imbalance Causes Objective Drift}
\vspace{- 0.05 in}
A number of studies explored detrimental effects of non-IID training data on the performance of a global model learned via FedAvg. An example is SCAFFOLD \citep{scaffold} which demonstrates \emph{objective drift} in non-IID FL manifested through large differences between local models $\mathbf{\theta}_{k}^{*}$ trained on substantially different data distributions. The drift is due to FedAvg updating the global model in the direction of the weighted average of local optimal models, which is not necessarily leading towards the optimal global model $\mathbf{\theta}^{*}$. The optimal model $\mathbf{\theta}^{*}$, in principle obtained by solving optimization in Eq. \ref{objective}, achieves minimal empirical error on the data with uniform label distribution and is intuitively closer to the local optimal models trained on balanced data. Recent work \citep{fedcbs} empirically verified this conjecture through extensive experiments. Let $\nabla F(\mathbf{\theta}^{t})$ denote the gradient of $F(\theta^{t})$ given the global model $\mathbf{\theta}^{t}$ at round $t$; the difference between $\nabla F(\mathbf{\theta}^{t})$ and the local gradient $\nabla F_{k}(\mathbf{\theta}^{t})$ computed on client $k$'s data is typically assumed to be bounded \citep{chen2020optimal, clustered, fednova}. To proceed, we formalize the assumption about the relationship between gradients and data label distributions.
\begin{assumption}[Bounded Dissimilarity.]
\label{bounded_disimilarity}
Gradient $\nabla F_{k}(\mathbf{\theta}^{t})$ of the $k$-th local model at global round $t$ is such that
\begin{equation}
\label{eq:3}
\begin{aligned}
    \left \Vert \nabla F_{k}(\mathbf{\theta}^{t}) - \nabla F(\mathbf{\theta}^{t})\right \Vert^{2} \leq \kappa -\rho e^{ \beta\left(H(\mathcal{D}^{(k)}) - H(\mathcal{D}_{0})\right)} = \sigma_{k}^{2} ,
\end{aligned}
\end{equation}
where $\mathcal{D}^{(k)}$ is the data label distribution of client $k$, $\mathcal{D}_{0}$ denotes uniform distribution, $H(\cdot)$ is Shannon's entropy of a stochastic vector, and $\beta > 0, \kappa > \rho > 0$.
\end{assumption}
The assumption commonly encountered in literature is recovered by setting the right-hand side of (\ref{eq:3}) to $\sigma^{2}_{m} = \max_{k} \sigma_{k}^{2}$.
Intuitively, if the data label distribution of client $k$ is highly imbalanced {(i.e., $H(\mathcal{D}^{(k)})$ is small), the local gradient $\nabla F_{k}(\mathbf{\theta}^{t})$ may significantly differ from the global gradient $ \nabla F(\mathbf{\theta}^{t})$ (as reflected by the bound above). 
 Analytically, connecting the gradients to the local data label distributions allows one to characterize the effects of client selection on the variance and the rate of convergence. The results of extensive experiments that empirically verify the above assumption are reported in Appendix \ref{empirical_validation}.
 
 \vspace{- 0.05 in}
\subsection{Estimating Client's Data Heterogeneity}
 \vspace{- 0.05 in}
If the server were given access to clients' data label distributions, selecting clients would be relatively straightforward \citep{haccs}. However, privacy concerns typically discourage clients from sharing such information. Previous studies have explored the use of multi-arm bandits for inferring clients' data heterogeneity from local model parameters, or have utilized a validation dataset at the server to accomplish the same \citep{efficientMAB, mab_distribution,fedcbs}. In this section, we demonstrate how to efficiently and accurately estimate data heterogeneity using local updates of the output layer of a neural network in a classification task. Figure \ref{fc layer} illustrates the last two layers in a typical neural network. The prediction $\mathbf{q} \in \mathbb{R}^{C}$ is computed by forming a weighted average of signals $\mathbf{z} \in \mathbb{R}^{L}$ utilizing the weight matrix $\mathbf{W}\in \mathbb{R}^{C\times L}$ and bias $\mathbf{b} \in \mathbb{R}^{C}$. 
\subsubsection{Local updates of the output layer}
\label{gradient_fc}
An empirical investigation of the gradients of the output layer's weights while training with FedAvg using mini-batch stochastic gradient descent (SGD) as an optimizer is reported in \citep{chen2024recovering,userlevel}. There, the focus is on detecting the presence of specific labels in a batch rather than on exploring the effects of class imbalance on the local update. To pursue the latter, we focus on the correlation between local updates of the output layer's bias and the client's data label distribution; we start by analyzing the training via FedAvg that employs SGD and then extend the results to other FL algorithms that utilize optimizers beyond SGD. We assume that the model is trained by minimizing the cross-entropy (CE) loss over one-hot labels -- a widely used multi-class classification framework. The gradient is computed by averaging contributions of the samples in mini-batches, i.e., $\nabla_{\mathbf{b}} \mathcal{L}_{\textbf{ce}} = \frac{1}{Bl}\sum_{j = 1}^{l}\sum_{n = 1}^{B} \nabla_{\mathbf{b}} \mathcal{L}^{(j,n)}_{\textbf{ce}}(\mathbf{x}^{(j,n)}, y^{(j,n)})$, where $B$ denotes the batch size, $l$ is the number of mini-batches, $\mathbf{x}^{(j,n)}$ is the $n$-th point in the $j$-th mini-batch and $y^{(j,n)} \in [C]$ is its label. The contribution of $\mathbf{x}^{(j,n)}$ to the $i$-th component of the gradient of the output layer's bias $\mathbf{b}$ can be found as (details provided in Appendix \ref{gradientoffc})
\begin{equation}
\label{grad_yn_paper}
    \nabla_{b_{i}} \mathcal{L}^{(j,n)}_{\textbf{ce}}(\mathbf{x}^{(j,n)}, y^{(j,n)}) = \mathbb{I}\{i = y^{(j,n)}\}\frac{-\sum_{c \not = i}\exp({q^{(j,n)}_{c}})}{\sum_{c = 1}^{C}\exp({q^{(j,n)}_{c}})} +  \mathbb{I}\{i \not = y^{(j,n)}\} \frac{\exp({q^{(j,n)}_{i}})}{\sum_{c = 1}^{C}\exp({q^{(j,n)}_{c}})}, 
\end{equation}
where $\mathbb{I}\{\cdot\}$ is an indicator, $\mathbf{q}^{(j,n)}  = \mathbf{W}\cdot\mathbf{z}^{(j,n)} + \mathbf{b}$ is the output logit for signals $\mathbf{z}^{(j,n)} \in \mathbb{R}^{L}$ corresponding to training point $(\mathbf{x}^{(j,n)},y^{(j,n)})$ (see Fig. \ref{fc layer}), and where $C$ denotes the number of classes. We make the following observations: (1) the sign of $y^{(j,n)}$-th component of
$\nabla_{\mathbf{b}} \mathcal{L}^{(j,n)}_{\textbf{ce}}$ is opposite of the sign of other components; and (2) the $y^{(j,n)}$-th component of $\nabla_{\mathbf{b}} \mathcal{L}^{(j,n)}_{\textbf{ce}}$ is equal in magnitude to all the other components combined. Note that the above two observations are standard for neural networks using CE loss for supervised multi-class classification tasks.



In each global round $t$ of FedAvg, the selected client $k$ starts from the global model $\theta^{t}$ and proceeds to compute local update in $R$ local epochs employing an SGD optimizer with learning rate $\eta$. According to Eq. \ref{grad_yn_paper}, the $i$-th component of local update $\Delta \mathbf{b}^{(k)}$ is computed as
\begin{equation}
\label{expectationofupdate}
    \Delta b^{(k)}_{i} = -\frac{\eta}{Bl}\sum^{l}_{j=1}\sum_{n=1}^{B}\sum_{r=1}^{R}  \nabla_{b_{i}} \mathcal{L}^{(j,n,r)}_{\textbf{ce}},
\end{equation}
where $\nabla_{b_{i}} \mathcal{L}^{(j,n,r)}_{\textbf{ce}}$ denotes the gradient of bias at local epoch $r$. Note that the local update of client $k$,  $\Delta \mathbf{b}^{(k)}$, is dependent on the label distribution of client $k$'s data, $\mathcal{D}^{(k)} = [D^{(k)}_{1}, \dots, D^{(k)}_{C}]^{T}$ and the label-specific components of $\mathbf{q}^{(j,n)}$ which change during training. We proceed by relating expected local updates to the  label distributions; for convenience, we first introduce the following definition.

\begin{definition}
\label{expection of output}
Let $\mathcal{B}^{-i}$ be the subset of local data $\mathcal{B}$ that excludes points with label $i$. Let $\textbf{s}^{-i}(\mathbf{x}) \in [0,1]^{C}$ be the softmax output of a trained neural network for a training point $(\mathbf{x}, y) \in \mathcal{B}^{-i}$. The $i$-th component of $\textbf{s}^{-i}(\mathbf{x})$, $\textbf{s}^{-i}_{i}(\mathbf{x})$, indicates the level of confidence in (erroneously) classifying $\textbf{x}$ as having label $i$.
For convenience, we define $
    \mathcal{E}_{i} = \mathbb{E}_{(\mathbf{x},y )\sim \mathcal{B}^{-i}}\left[ \mathbf{s}^{-i}_{i}(\mathbf{x})\right], \forall i \in [C]$.

\end{definition}

In an untrained/initialized neural network where classifier makes random predictions, $\mathcal{E}_{i} = 1/C$; as training proceeds, $\mathcal{E}_{i}$ decreases. By taking expectation and simplifying, we obtain (details provided in Appendix \ref{AnalysisOfexpection})
\begin{equation}
\label{expactation_b}
    \mathbb{E}\left[\Delta b^{(k)}_{i}\right] =  \eta R \left(D_{i}^{(k)}\sum_{c=1}^{C}\mathcal{E}_{c} -\mathcal{E}_{i}\right), 
\end{equation}
where $D_{i}^{(k)}$ denotes the true fraction of samples with label $i$ in client $k$'s data, $\sum_{i = 1}^{C}D_{i}^{(k)} = 1$.  

\subsubsection{Estimating local data heterogeneity}

We quantify the heterogeneity of clients' data by an entropy-like measure defined below. Let $\mathcal{D}^{(k)}$ denote the label distribution of client $k$'s data; its entropy is defined as $H(\mathcal{D}^{(k)}) \triangleq -\sum_{i=1}^{C}D_{i}^{(k)} \ln D_{i}^{(k)} \leq \ln C$. Recall that more balanced data results in higher entropy, and that $H(\mathcal{D}^{(k)})$ takes the maximal value when $\mathcal{D}^{(k)}$ is uniform. The server does not know $\mathcal{D}^{(k)}$ and therefore cannot compute $H(\mathcal{D}^{(k)})$ directly. We define
\begin{equation}
    \hat{H}(\mathcal{D}^{(k)}) \triangleq H(\mbox{softmax}(\Delta\mathbf{b}^{(k)}, T)),
\end{equation}
here $T$ is a scaling hyper-parameter (so-called \emph{temperature}). 
Note that even though we can compute $\hat{H}(\mathcal{D}^{(k)})$ to characterize heterogeneity, $D_{i}^{(k)}$ and $\mathcal{E}_{i}$ remain unknown to the server (details in \ref{privacy}).

\begin{theorem}
\label{theorem_estimation}
Consider an FL system in which clients collaboratively train a model for a classification task over $C$ classes. Let $\mathcal{D}^{(u)}$ and $\mathcal{D}^{(k)}$ denote data label distributions of an arbitrary pair of clients $u$ and $k$, respectively. Moreover, let \textbf{U} denote the uniform distribution, and let $\eta$ and $R$ be the learning rate and the number of local epochs, respectively. Then
\begin{equation}
\label{theorem1}
   \mathbb{E}\left[ \hat{H}(\mathcal{D}^{(u)}) - \hat{H}(\mathcal{D}^{(k)}) \right] \geq  \frac{1}{2}  \left(\frac{ \eta R  }{CT}\sum_{c=1}^{C} \mathcal{E}_{c} \right)^{2} \left\Vert \mathcal{D}^{(k)} -\mathbf{U}\right\Vert_{2}^{2} 
    -\frac{ \eta R }{T} \left\Vert  \mathcal{D}^{(u)} - \mathbf{U}\right\Vert_{\infty} - \mathcal{C}\delta,      
\end{equation}
where $\mathcal{C} = \frac{ \eta R (\eta R + C^{2}T\ln C) }{C^{2}T^{2}}$ and $\delta = \max_{i} \left|\frac{\sum_{c=1}^{C}\mathcal{E}_{c}}{C} - \mathcal{E}_{i}\right|$. The proof is provided in Appendix \ref{proof_theorem_1}.
\end{theorem}
 As an illustration, consider the scenario where client $u$ has a balanced dataset while the dataset of client $k$ is imbalanced; then $\|\mathcal{D}^{(k)} - \mathbf{U}\|_{2}^{2}$ is relatively large compared to $\|\mathcal{D}^{(u)} - \mathbf{U} \|_{\infty}$. The bound in (\ref{theorem1}) also depends on $\delta$, which is reflective of how misleading on average can a class be; small $\delta$ suggests that no class is universally misleading. As shown in Appendix \ref{AnalysisOfexpection}, during training $\delta$ gradually decreases to $0$ as $\sum_{i=1}^{C} \mathcal{E}_{i}$ decreases to $0$.
\subsubsection{Generalizing beyond FedAvg and SGD}
The proposed method for estimating clients' data heterogeneity relies on the properties of the gradient for the cross-entropy loss objective discussed in Section \ref{gradient_fc}. However, for FL algorithms other than FedAvg, such as FedProx \citep{fedprox}, FedDyn \citep{feddyn} and Moon \citep{moon}, which add regularization to combat overfitting, the aforementioned properties may not hold. Moreover, optimization algorithms using second-order momentum such as Adam \citep{adam} deploy update rules different from SGD, making the local updates no longer proportional to the gradients. Nevertheless, HiCS-FL remains capable of distinguishing between clients with imbalanced and balanced data, which will be demonstrated in our experiments. Further theoretical discussion of various FL algorithms with optimizers beyond SGD are in appendix \ref{regularization_appendix} and \ref{optimizers_appendix}.

\vspace{- 0.05 in}
\subsection{Heterogeneity-guided Clustering}
\vspace{- 0.05 in}
\label{sampling}
Clustered Sampling \citep{clustered} uses cosine similarity \citep{sattler2020clustered} between gradients to quantify proximity between clients' data distributions and subsequently group them into clusters. However, cosine similarity cannot help distinguish between clients with balanced and those with imbalanced datasets. Motivated by this observation, we introduce a new distance measure that incorporates estimates of data heterogeneity $\hat{H}(\mathcal{D}^{(k)})$. In particular, the proposed measure of distance between clients $u$ and $k$ that we use to form clusters is defined as
\begin{equation}
\label{clustering}
     \textbf{Distance}(u,k) =  \text{arc}\cos\left(
    \frac{\Delta \mathbf{b}^{(u)}\cdot \Delta \mathbf{b}^{(k)} }{|\Delta \mathbf{b}^{(u)}|\cdot|\Delta \mathbf{b}^{(k)}|}\right)  + \lambda \left| \hat{H}(\mathcal{D}^{(u)}) - \hat{H}(\mathcal{D}^{(k)})\right|,
\end{equation}
where the first term is akin to the cosine similarity used by CS with the major difference that we compute it using only the updates of the bias in the output layer, which is much more efficient than using the weights of the entire network; $\lambda$ is a pre-defined hyper-parameter (set to $10$ in all our experiments). For large $\lambda$, the second term dominates when there are clients with different levels of statistical heterogeneity; this allows emergence of clusters that group together clients with balanced datasets. The second term is small when clients have data with similar levels of statistical heterogeneity; in that case, the distance measure reduces to the conventional cosine similarity.

\vspace{- 0.05 in}
\subsection{Hierarchical Clustered Sampling}
\label{sampling_policy}
To select $K$ out of $N$ clients in an FL system, we first organize
the clients into $M \geq K$ groups via the proposed Hierarchical Clustered Sampling (HiCS) technique. In particular, during the first $\lceil N/K\rceil$ training rounds the server randomly (without replacement) selects clients and collects from them local updates of $\Delta \mathbf{b}^{(k)}$; the server then estimates $\hat{H}^{t}(\mathcal{D}^{(k)})$ for each selected client $k$ and clusters the clients using the distance measure defined in Eq. \ref{clustering}. Let $G_{1}^{t}, \dots, G_{M}^{t}$ denote the resulting $M$ clusters at global round $t$, and let $\bar{H}_{m}^{t} = \frac{1}{|G_{m}|}\sum_{k \in G_{m}} \hat{H}^{t}(\mathcal{D}^{(k)})$ characterize the average heterogeneity of clients in cluster $m$, $m \in [M]$. Having computed $\bar{H}_{m}^{t}$, HiCS selects a cluster according to the probability vector $\mathbf{\pi}^{t}$, and then from the selected cluster selects a client according to the probability vector $\Tilde{\mathbf{p}}_{m}^{t}$. The two probability vectors  $\mathbf{\pi}^{t}$ and $\Tilde{\mathbf{p}}_{m}^{t}$ are defined as
\begin{equation}
\label{policy}
    \mathbf{\pi}^{t} = \left[ \frac{\exp(\gamma^{t} \bar{H}_{1}^{t})}{\sum_{m = 1}^{M}\exp(\gamma^{t} \bar{H}_{m}^{t})}, \dots,  \frac{\exp(\gamma^{t} \bar{H}_{M}^{t})}{\sum_{m = 1}^{M}\exp(\gamma^{t} \bar{H}_{m}^{t})} \right], 
    \Tilde{\mathbf{p}}_{m}^{t} = \left[ \frac{p_{k_{1}}}{\sum_{k \in G_{m}}p_{k}}, \dots,  \frac{p_{k_{|G_{m}|}}}{\sum_{k \in G_{m}}p_{k}}\right], 
\end{equation}
where $k_{1}, \dots, k_{|G_{m}|}$ are the indices of clients in cluster $G_{m}$, $\gamma^{t} = \gamma^{0}( 1 - \frac{t}{\mathcal{T}})$ denotes an annealing hyper-parameter, and $\mathcal{T}$ is the number of global rounds. The annealing parameter is scheduled such that at first it promotes sampling clients with balanced data, thus accelerating and stabilizing the convergence of the global model. To avoid overfitting potentially caused by repeatedly selecting a small subset of clients, the annealing parameter is gradually reduced to $\gamma^{t} \approx  0$, when the server samples the clusters uniformly. The described procedure is formalized as Algorithm \ref{alg:fedhics}.

\begin{algorithm}[t]
\vspace{-0.15 in}
\caption{HiCS-FL}
\label{alg:fedhics}
\begin{multicols}{2}
\begin{algorithmic}[1] 
\small
\REQUIRE ~~\\ %
    Datasets distributed across $N$ clients, the number of clients to sample $K$, total global rounds $\mathcal{T}$.
    \STATE Initialize updates of bias $\Delta \mathbf{b}^{(k)} \leftarrow \mathbf{0}\; \forall k \in [N]$, global model $\mathbf{\theta}^{t} \leftarrow \mathbf{\theta}^{1}$, $S_{0} = [N].$
     \FOR{$t = 1,\dots,\mathcal{T}$}
    \IF{$t \leq \lceil N/K \rceil$}
    \STATE  $\mathcal{S}^{t}  \leftarrow$ randomly sample $\min (K,|S_{0}|)$\\ clients from $S_{0}$, update $S_{0} \leftarrow S_{0} - \mathcal{S}^{t}$;
    \ELSE
    \STATE estimate $\hat{H}^{t}(\mathcal{D}^{(k)})$ and cluster $N$ clients into $M$ groups based on Eq. \ref{clustering};
    \STATE  $\mathcal{S}^{t}  \leftarrow \emptyset$;
    \WHILE{$|\mathcal{S}^{t}| < K$}
    \STATE  sample group $G_{m}^{t}$ according to  $\pi^{t}$;
    \STATE sample client $k$ in $G_{m}^{t}$ based on  $\Tilde{\mathbf{p}}_{m}$;
    \STATE  $\mathcal{S}^{t}  \leftarrow \mathcal{S}^{t} \cup k$;
    \ENDWHILE
    \ENDIF
    \FOR{$k \in \mathcal{S}^{t}$}
    \STATE $\mathbf{\theta}_{k}^{t}\leftarrow$ \textbf{LocalUpdate}$(\mathbf{\theta}^{t})$, $\Delta\mathbf{b}^{(k)} \in \mathbf{\theta}_{k}^{t} - \mathbf{\theta}^{t}$ 
    \ENDFOR
    \STATE  $\mathbf{\theta}^{t+1} \leftarrow \frac{1}{K}\sum_{k \in \mathbf{S}^{t}} \mathbf{\theta}_{k}^{t}$;
    \STATE $\Delta\mathbf{b}^{(k)} \leftarrow \Delta \mathbf{b}^{(k)}, \forall k \in \mathcal{S}^{t}$;
    \ENDFOR
\ENSURE ~~\\ 
    The global model $\mathbf{\theta}^{\mathcal{T}+1}$
\end{algorithmic}
\end{multicols}
\vspace{-0.1 in}
\end{algorithm}

\vspace{- 0.05 in}
\subsection{Convergence Analysis}
\label{convergence}
Adopting the standard assumptions of smoothness, unbiased gradients and bounded variance \citep{chen2020optimal}, the following theorem holds for FedAvg with SGD optimizer.
\begin{theorem}
\label{theorem 2}
Assume $F_{k}(\cdot)$ is $L$-smooth for all $k \in [N]$. Let $\theta^{t}$ denote parameters of the global model and let $F(\cdot)$ be defined as in Eq. \ref{objective}. Furthermore, assume the stochastic gradient estimator $g_{k}(\theta^{t})$ is unbiased and the variance is bounded such that $\mathbb{E}\left \Vert g_{k}(\theta^{t}) - \nabla F_{k}(\theta^{t})\right \Vert^{2} \leq \sigma^{2}$. Let $\eta$ and $R$ be the learning rate and the number of local epochs, respectively. If the learning rate is such that $\eta \leq \frac{1}{8LR}, R \geq 2$, then
\begin{equation}
\begin{aligned}
 \min_{t \in [\mathcal{T}]} \left\Vert\nabla F(\theta^{t})\right\Vert^{2} \leq \frac{1}{ \mathcal{T}}\left(\frac{F(\theta^{0}) - F(\theta^{*})}{\mathcal{A}_{1}} + \mathcal{A}_{2}\sum_{t=0}^{\mathcal{T}-1}\sum_{k=1}^{N}\omega_{k}^{t}\sigma_{k}^{2}\right)  + \mathbf{\Phi},
\end{aligned}
\end{equation}
where $\mathcal{A}_{1}$, $\mathcal{A}_{2}$, $\mathbf{\Phi}$ are positive constants, and $\omega_{k}^{t}$ is the probability of sampling client $k$ at round $t$.
\end{theorem}
Note that only the second term in the parenthesis on the right-hand side of the bound 
in Theorem \ref{theorem 2} is related to the sampling method $\Pi$. Under Assumption \ref{bounded_disimilarity},
\begin{equation}
\sum_{k=1}^{N}\omega_{k}^{t}\sigma_{k}^{2} \leq  \kappa - \sum_{k=1}^{N}\omega_{k}^{t}\frac{\exp\left(\beta H(\mathcal{D}^{(k)})\right)}{\exp\left(\beta H(\mathcal{D}_{0})\right)}\rho  = \kappa -\mathcal{H}_{\Pi} .
\end{equation}
If the server samples clients with weights proportional to $p_{k}$, 
the statistical heterogeneity of the entire FL system may be characterized by
$\mathcal{H}_{\mathbf{S}} = \sum_{k=1}^{N} p_{k} \frac{\exp(\beta(H(\mathcal{D}^{(k)}))}{\exp(\beta(H(\mathcal{D}_{0}))}\rho$. If all clients have class-imbalanced data, $\mathcal{H}_{\mathbf{S}}$ is small and thus random sampling leads to unsatisfactory convergence rate (as indicated by Theorem~\ref{theorem 2}). On the other hand, since the clients sharing a cluster have similar data entropy, the proposed HiCS-FL leads to $\omega_{k}^{t} = \frac{ p_{k}\exp(\gamma^{t}\hat{H}^{t}(\mathcal{D}^{(k)}))}{\sum_{j=1}^{N}p_{j}\exp(\gamma^{t}\hat{H}^{t}(\mathcal{D}^{(j)}))}$. When training starts, $\mathcal{H}_{\Pi}$ is large because the server tends to sample clients with higher $p_{k}\exp(\gamma^{t}H(\mathcal{D}^{(k)}))$; as $\gamma^{t}$ decreases, $\mathcal{H}_{\Pi}$ eventually approaches $\mathcal{H}_{\mathbf{S}}$. Further details and the proof of the theorem are in Appendix \ref{converge}.

 \vspace{- 0.1 in}
\section{Experiments}
 \vspace{- 0.05 in}
\label{exp}
\textbf{Setup.} We evaluate the proposed HiCS-FL algorithm on four benchmark datasets (FMNIST, CIFAR10, Mini-ImageNet and THUC news) using different model architectures. We use four baselines: random sampling, pow-d \citep{powerofchoice}, clustered sampling (CS) \citep{clustered}, DivFL \citep{diverse} and FedCor \citep{fedcor}. To generate non-IID data partitions, we follow the strategy in \citep{yurochkin2019bayesian}, utilizing Dirichlet distribution with different concentration parameters $\alpha$ which controls the level of heterogeneity (smaller $\alpha$ leads to generating less balanced data). In a departure from previous works we utilize several different $\alpha$ to generate data partitions for a single experiment, leading to a realistic scenario of varied data heterogeneity across different clients. To quantify the performance of the tested methods, we use two metrics: (1) average training loss, and (2) test accuracy of the learned global model. 
For better visualization, data points in the results are smoothened by a Savitzky–Golay filter with window length $13$ and the polynomial order set to $3$. Further details of the experimental setting and a visualization of data partitions are in Appendix \ref{experimental_settings} and \ref{visualization_appendix}.
 
 \vspace{- 0.05 in}
\subsection{Comparison on Test Accuracy and Training Loss}
\begin{figure*}[t] 
    \centering
	  \subfloat[FMNIST (1)]{
       \includegraphics[width=0.25\linewidth]{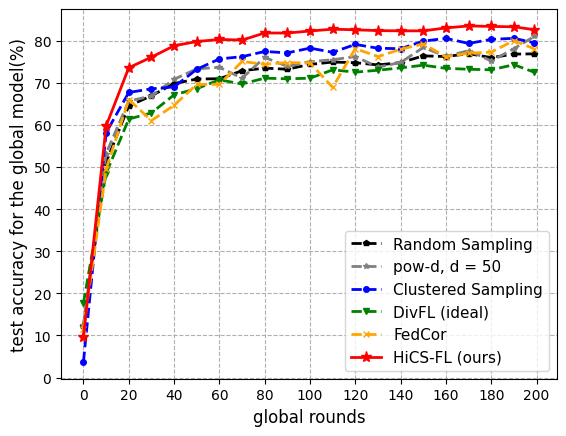}}
              \hspace{ 0.2in}
	  \subfloat[FMNIST (2)]{
        \includegraphics[width=0.25\linewidth]{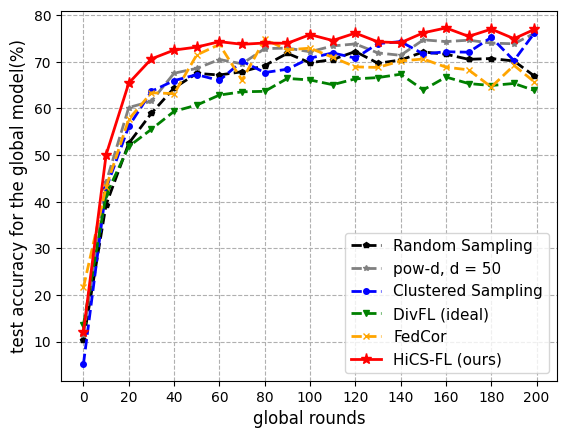}}
               \hspace{ 0.2in}
	  \subfloat[FMNIST (3)]{
        \includegraphics[width=0.25\linewidth]{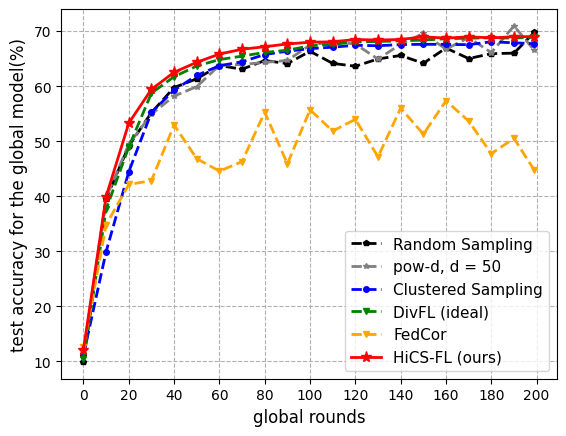}}
        \quad
        \vspace{- 0.1 in}

         \subfloat[CIFAR10 (1)]{
       \includegraphics[width=0.25\linewidth]
       {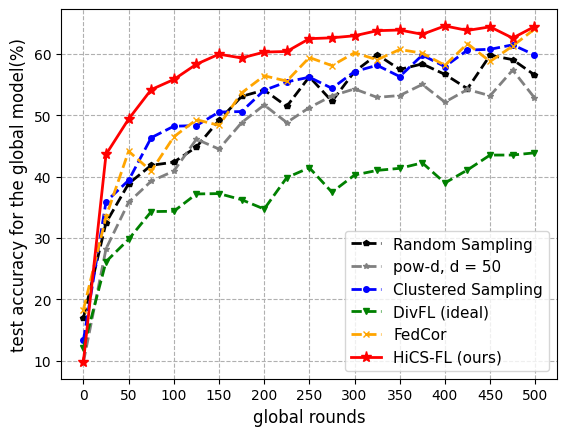}}
              \hspace{ 0.2in}
	  \subfloat[CIFAR10 (2)]{
        \includegraphics[width=0.25\linewidth]{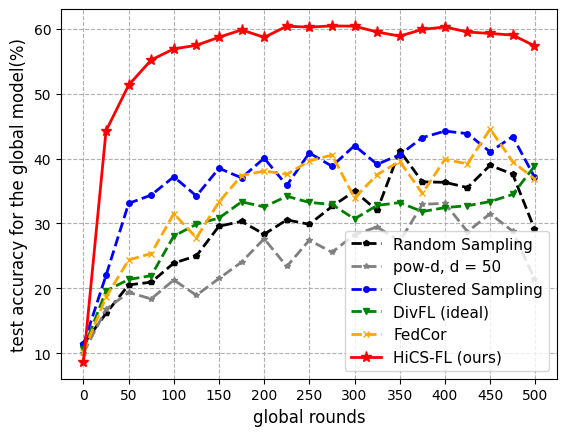}}
               \hspace{ 0.2in}
	  \subfloat[CIFAR10 (3)]{
        \includegraphics[width=0.25\linewidth]{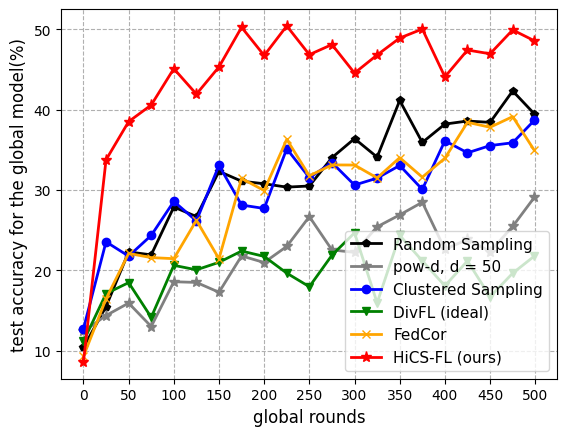}}
        
	\caption{Test accuracy for the global model on 3 groups of data partitions of FMNIST and CIFAR10.}
\label{FMNIST_CIFAR10_accuracy} 
 \vspace{- 0.2 in}
\end{figure*}

\textbf{FMNIST.} We run FedAvg with \textbf{SGD} to train a global model which has CNN architecture in an FL system with $50$ clients, where $10\%$ of clients are selected to participate in each round of training. 
The data partitions are generated using one of $3$ sets of the concentration parameter $\alpha$ values: (1) $\{0.001,0.002,0.005, 0.01,0.5\}$; (2) $\{0.001,0.002,0.005, 0.01,0.2\}$; (3) $\{0.001\}$. These are used to generate clients' data so as to emulate the following scenarios: (1) $80\%$ of clients have severely imbalanced data while the remaining $20\%$ have balanced data; (2) $80\%$ clients have severely imbalanced data while the remaining $20\%$ have mildly imbalanced data; (3) all clients have severely imbalanced data. Note that $\mathcal{H}_{\mathbf{M}}$ monotonically decreases as we go through settings (1) to (3). For a fair comparison, pow-d and DivFL are deployed with their ideal settings where the server requires all clients to precompute in each round a metric that is then used for client selection. Figure \ref{FMNIST_CIFAR10_accuracy} shows that HiCS-FL outperforms other methods across different settings, exhibiting the fastest convergence rates and the least amount of variance. Particularly significant is the acceleration of convergence in setting (1) where $20\%$ of the participating clients have balanced data. 
Figure \ref{loss} shows that HiCS-FL is helping achieve significant reduction of training variations (as expected, see Section \ref{convergence}) as evident by a smooth loss trajectory.

\textbf{CIFAR10.} Here we  compare the performance of HiCS-FL to FedProx \citep{fedprox} running CNN model with \textbf{Adam} optimizer on the task of training an FL system with 50 clients, where $20\%$ of clients are selected to participate in each training round. Similar to the experiments on FMNIST, $3$ sets of the concentration parameter $\alpha$ are considered: (1) $\{0.001, 0.01, 0.1, 0.5, 1\}$; (2) \{$0.001$, $0.002$, $0.005$, $0.01$, $0.5$\}; (3) $\{0.001, 0.002, 0.005, 0.01, 0.1\}$. The interpretation of the scenarios emulated by these setting is as same as in the FMNIST experiments. Figure \ref{FMNIST_CIFAR10_accuracy} demonstrates improvement of HiCS-FL over all the other methods. HiCS-FL exhibits particularly significant improvements in settings (2) and (3), where $80\%$ of the clients with extremely imbalanced data benefit from $20\%$ of the clients with either balanced or mildly imbalanced data. 
The advantage of HiCS-FL in setting (1) where all clients have relatively high data heterogeneity is relatively modest (see Fig.\ref{FMNIST_CIFAR10_accuracy}.(d)) because the system's $\mathcal{H}_{\mathbf{S}}$ is relatively large (see discussion in Section \ref{convergence}). 
\begin{figure*}[t] 
    \centering
	  \subfloat[FMNIST (1)]{
       \includegraphics[width=0.25\linewidth]
       {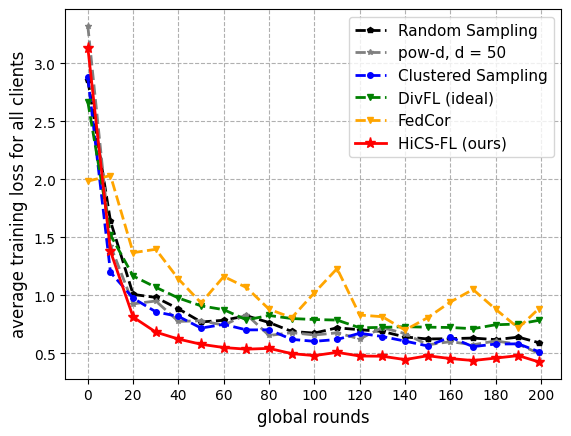}}
       \hspace{ 0.2in}
	  \subfloat[CIFAR10 (1)]{
        \includegraphics[width=0.25\linewidth]{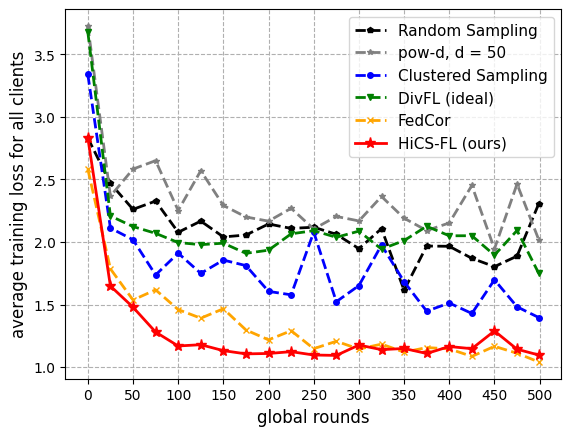}}
        \hspace{ 0.2in}
        \subfloat[Mini-ImageNet (1)]{
        \includegraphics[width=0.25\linewidth]{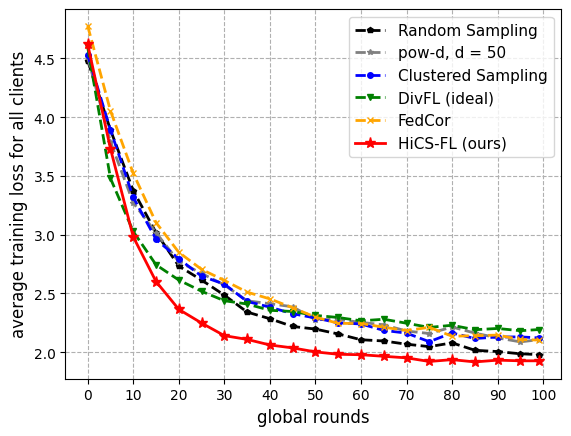}}
	\caption{Training loss of HiCS-FL compared to four baselines for setting (1) on the three datasets.}
\label{loss} 
 \vspace{- 0.1 in}
\end{figure*}

\textbf{Mini-ImageNet.} As in the Mini-ImageNet experiments, we compare HiCS-FL to FedProx running ResNet18 with \textbf{Adam} optimizer but now consider training of an FL system with 100 clients, where $20\%$ of the clients are selected to participate in each round of training. We consider two settings of the concentration parameter $\alpha$: (1) $\{ 0.001,0.01,0.1,0.5,1 \}$ and (2) $\{ 0.001,0.005,0.01,0.1,1 \}$. Setting (1) emulates the scenario where clients have a range of heterogeneity profiles, from extremely imbalanced, through mildly imbalanced, to balanced, while setting (2) corresponds to the scenario where $80\%$ of the clients have extremely imbalanced data while the remaining $20\%$ have balanced data. The system's $\mathcal{H}_{\mathbf{S}}^{(1)}$ for setting (1) is larger than $\mathcal{H}_{\mathbf{S}}^{(2)}$ for setting (2), which is reflected in a more significant improvements achieved by HiCS-FL in the latter setting, as shown in Figure \ref{tiny_acc}.

\textbf{THUC news.} To evaluate our method on data from a different domain, we conduct experiments involving text classification on the \href{https://github.com/649453932/Chinese-Text-Classification-Pytorch}{THUC news} dataset in Chinese language ($10$ labels). Similar to the aforementioned experiments, we allocate data to $50$ clients by emulating heterogeneous data distributions scenarios with parameter $\alpha$ set to: (1) \{$0.001$, $0.01$, $0.1$, $0.2$,$1$\}; (2) $\{0.001, 0.002, 0.01, 0.1,0.5\}$; and (3) \{$0.001$, $0.002$, $0.005$, $0.01$, $0.1$\}. We trained TextRNNs \citep{textrnn} with BiLSTM architecture as the classifiers using \textbf{Adam} optimizer. The test accuracy of the global model trained with different schemes for $100$ global rounds, reported in Table~\ref{table1}, show that our method outperforms baselines in all the settings, demonstrating efficacy of our proposed algorithm in a simple NLP task.


 

\begin{table}[]
\caption{Test accuracy (\%) for the global model on 3 groups of data partitions of THUC news dataset. \\}
\centering
\begin{tabular}{lcccccc}
\bottomrule[1pt]
\label{table1}
Schemes & Random       & Pow-of-Choice  & CS  &DivFL &FedCor & \textbf{HiCS-FL} \\
\hline
\rowcolor[HTML]{EFEFEF} 

settng (1)           &  78.9                         &80.0                            &80.6 &73.0 &81.2 &\textbf{83.2}   \\
settng (2)            & 74.9                             &75.4                              &  82.8 &68.9 &81.3 & \textbf{83.9}                 \\
\rowcolor[HTML]{EFEFEF}
settng (3)          & 72.7                             &66.5                             &79.4 &72.1 &76.4 &\textbf{79.7}     \\
\toprule[1pt]
\end{tabular}
\end{table}
\begin{table*}[t]
\caption{The number of communication rounds needed to reach a certain test accuracy in the experiments on FMNIST, CIFAR10, Mini-ImageNet and THUC News. All results are for the  concentration parameter setting (2).}
\small
\begin{center}
\vspace{-0.1 in}
\label{table2}
\begin{tabular}{ccccccccc}
\bottomrule[1pt]
\multirow{2}{*}{Schemes} & \multicolumn{2}{c}{FMNIST}  &\multicolumn{2}{c}{CIFAR10}  &\multicolumn{2}{c}{Mini-ImageNet} & \multicolumn{2}{c}{THUC news}  \\
\cline{2-9}
  & \text{acc} = 0.75 &speedup
 & \text{acc} = 0.6 &speedup  & \text{acc} = 0.5  & speedup  & \text{acc} = 0.8  & speedup \\
\hline
\rowcolor[HTML]{EFEFEF} 
Random     
 &  149& 1.0$\times$&  898 & 1.0$\times$  &  191 & 1.0$\times$ &83 & 1.0$\times$ 
  \\
pow-d  & 79&1.8$\uparrow$ & 1037& 0.9$\downarrow$ &432  & 0.4$\downarrow$&109 & 0.8$\downarrow$ \\
\rowcolor[HTML]{EFEFEF} 
CS      & 114& 1.3$\uparrow$ &  748 & 1.2$\uparrow$  &186  &1.0$\times$  &74 &1.1$\uparrow$ 
 \\
DivFL   & 478 & 0.3$\downarrow$  & 1417 & 0.6$\downarrow$ &726   &0.3 $\downarrow$ &289 &0.3$\downarrow$ \\
\rowcolor[HTML]{EFEFEF} 
FedCor   &88  &1.7$\uparrow$ &711  & 1.3$\uparrow$ &229   &0.8$\uparrow$  &100 &0.8$\downarrow$ \\
\textbf{HiCS-FL} & \textbf{60}   & \textbf{2.5$\uparrow$}&\textbf{123}& \textbf{7.3$\uparrow$} &\textbf{86}  &\textbf{2.2$\uparrow$} &\textbf{27} &\textbf{3.1$\uparrow$} \\
\toprule[1pt]
\end{tabular}
\end{center}
\vspace{-0.15in}
\end{table*}

\vspace{- 0.05 in}
\subsection{Accelerating the Training Convergence}
\begin{wrapfigure}[15]{r}{4cm}
\centering
\vspace{-0.15in}
\includegraphics[width=0.28\textwidth]{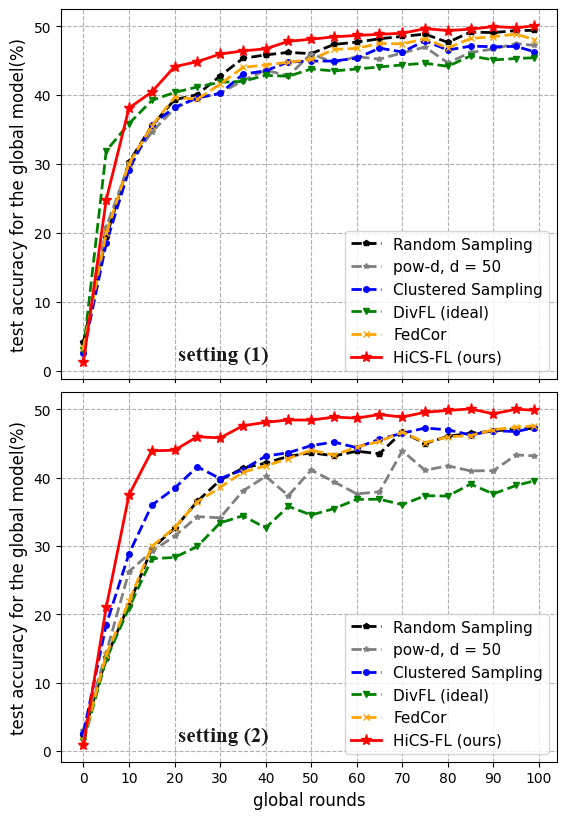}
\vspace{-0.1in}
\caption{MiniImageNet acc.}
\label{tiny_acc}
\end{wrapfigure}
In this section we report the communication costs required to achieve convergence when using HiCS-FL, and compare those results with the competing schemes. For brevity, we select one result from each experiment conducted on the considered four datasets, and display them in Table \ref{table2}. As can be seen from the table, HiCS-FL significantly reduces the number of communication rounds needed to reach target test accuracy. On FMNIST, HiCS-FL needs $60$ rounds to reach test accuracy $0.75$, achieving it $2.5$ times faster than the random sampling scheme. On CIFAR10, HiCS-FL requires only $123$ rounds to reach $0.6$ test accuracy, which is $7.3$ times faster than random sampling. Significant speedup appears on THUC dataset, in which HiCS-FL only needs $27$ rounds to achieve $0.8$ test accuracy, $3.1$ times faster than the baseline. Acceleration on Mini-ImageNet is relatively modest but HiCS-FL still outperforms other methods, and does so up to $2.2$ times faster than random sampling. 

Table~\ref{table2} also shows that HiCS-FL provides the reported improvements without introducing major computational and communication overhead. The only additional computation is due to estimating data heterogeneity and performing clustering utilizing bias updates, which scales with the total number of classes but does not increase with the size of the neural network model $\left|\theta^{t}\right|$. Remarkably, HiCS-FL outperforms pow-d, Clustered Sampling, DivFL and FedCor in terms of convergence speed, variance and test accuracy while requiring significantly less computations. More details are provided in Appendix \ref{complexity}.

\vspace{- 0.1 in}
\section{Conclusion}
\vspace{- 0.1 in}
In this paper, we studied federated learning systems where clients that own non-IID data collaboratively train a global model; the system operates under communication constraints and thus only a fraction of clients participates in any given round of training. 
We developed HiCS-FL, a hierarchical clustered sampling method which estimates clients' data heterogeneity and uses this information to cluster and select clients to participate in training. We analyzed the performance of the proposed heterogeneity estimation method, and the convergence of training a FL system that deploys HiCS-FL. Extensive benchmarking experiments on four datasets demonstrated significant benefits of the proposed method, including improvement in convergence speed, variance and test accuracy, accomplished with only a minor computational overhead.



{\small
\bibliographystyle{neurips_2024}
\bibliography{neurips_2024}
}

\newpage
\appendix

\section{Appendix}
\subsection{Details of the Experiments}
\label{experimental_settings}
\subsubsection{General Settings}
The experimental results were obtained using Pytorch \citep{paszke2019pytorch}. In the experiments involving FMNIST, each client used a CNN-based classifier with two $5\times5$-convolutional layers and two $2\times2$-maxpooling layers (with a stride of $2$), followed by a fully-connected layer. In the experiments involving CIFAR10, each client used a CNN-based classifier with three $3\times3$-convolutional layers and two $2\times2$-maxpooling layers (with a stride of $2$), followed by two fully-connected layers; dimension of the hidden layer was $64$. In the experiments involving Mini-ImageNet and THUC news, each client fine-tuned a pretrained \href{https://pytorch.org/vision/main/models/generated/torchvision.models.resnet18.html}{ResNet18} \citep{resnet} and learned a TextRNNs \citep{textrnn}, respectively.
The optimizers used for model training in the experiments on FMNIST and CIFAR10/Mini-ImageNet/THUC news were the mini-batch stochastic gradient descent (SGD) and Adam \citep{adam}, respectively. The learning rate was initially set to $0.001$ and then decreased every 10 iterations, with a decay factor $0.5$. The number of global communication rounds was set to 200, 500, 100 and 100 for the experiments on FMNIST, CIFAR10, Mini-ImageNet and THUC news, respectively. In all the experiments, the number of local epochs $R$ was set to $2$ and the size of a mini-batch was set to $64$. The sampling rate (fraction of the clients participating in a training round) was set to $0.1$ for the experiments on FMNIST/THUC news, and to $0.2$ for the experiments on CIFAR10/Mini-ImageNet. For the sake of visualization, data points in the presented graphs were smoothened by a Savitzky–Golay filter \citep{schafer2011savitzky} with window length $13$ and the polynomial order set to $3$.

 \subsubsection{Hyper-parameters}
 In all experiments, the hyper-parameter $\mu$ of the regularization term in FedProx \citep{fedprox} was set to $0.1$. In the Power-of-Choice (pow-d) \citep{powerofchoice} selection strategy, $d$ was set to the total number of clients: $50$ in the experiments on FMNIST, CIFAR10 and THUC news, $100$ in the experiments on Mini-ImageNet. When running DivFL \citep{diverse}, we used the ideal setting where 1-step gradients were requested from all client in each round (regardless of their participation status), similar to the Power-of-Choice settings. For FedCor \citep{fedcor}, we followed all settings in the paper and set the annealing coefficient $\beta$ controlling the sampling strategy to 0.9 as suggested in the paper. For HiCS-FL (our method), the scaling parameter $T$ (temperature) used in data heterogeneity estimation was set to $0.0025$ in the experiments on FMNIST and to $0.0015$ in the experiments on CIFAR10/Mini-ImageNet. In all experiments, parameter $\lambda$ which multiplies the difference between clients' estimated data heterogeneity (used in clustering) was set to $10$. In all experiments, the number of clusters $m$ was for convenience set to be equal to the number of selected clients $K$. The coefficient $\gamma^{0}$ was set to $4$ in the experiments on FMNIST and CIFAR10 while set to $2$ in the experiments on Mini-ImageNet. To group clients, both Clustered Sampling \citep{clustered} and HiCS-FL (our method) utilized an off-the-shelf clustering algorithm performing hierarchical clustering with Ward's Method.

\subsection{Empirical Validation of Assumption \ref{bounded_disimilarity}}
\label{empirical_validation}
To illustrate and empirically validate Assumption \ref{bounded_disimilarity}, we conducted extensive experiments on FMNIST and CIFAR10 with the same model mentioned in Section \ref{experimental_settings}. In particular, we varied $\alpha$ over $250$ values in the interval $[0.01,50]$ to generate data partitions allocated to $250$ clients; entropy of the generated label distributions ranged from $0$ to $\ln 10$ (maximum). In these experiments, we allowed all clients to participate in each of $500$ training rounds. To facilitate the desired study, in addition to these 250 clients we also simulated a super-client which owns a data set aggregating the data from all the clients (the set of labels in the aggregated dataset is uniformly distributed). In each round, clients start from the initialized global model and compute local gradients on their datasets; the super-client does the same on the aggregated dataset. The server computes and records squared Euclidean norm of the difference between the local gradients and the ``true" gradient (i.e., the super-client's gradient). In each round, the difference between the local gradient and the true gradient changes in a pattern similar to what is stated in Assumption \ref{bounded_disimilarity}. As an illustration, we plot all such gradient differences computed during the entire training process of a client. Specifically, the server computes the difference between local gradient and the true gradient in each round of training, obtaining $250\times500 = 12500$ data points that correspond to $250$ data partitions. For better visualization, we merged adjacent points.

The results obtained by following these steps in experiments on FMNIST and CIFAR10 are shown in Figure~\ref{ASSUMPTION}. For a more informative visualization, the horizontal coordinate of a point in the scatter plot is $ H(\mathcal{D}^{(k)})$, while the vertical coordinate is $\left \Vert \eta_{t} \nabla F_{k}(\mathbf{\theta}^{t}) -  \eta_{t} \nabla F(\mathbf{\theta}^{t})\right \Vert^{2}$. The dashed lines correspond to the curves $y = -\exp(\beta \left[ x - H(\mathcal{D}_{0})\right])\rho  + \kappa$ that envelop the majority of the generated points. In the case of FMNIST, the blue dashed line is parametrized by $\beta = 1.0$, $\rho = 0.13$, and $\kappa = 0.14$ while the green dashed line is parametrized by $\beta = 1.5$, $\rho = 0.025$, and $\kappa = 0.022$; these two lines envelop $95\%$ of the generated points. In the case of CIFAR10, the blue dashed line is parametrized by $\beta = 2.0$, $\rho = 0.30$, and $\kappa = 0.36$ while the green dashed line is parametrized by $\beta = 1.8$, $\rho = 0.15$, and $\kappa = 0.20$; as in the other plot, these two lines envelop $95\%$ of the generated points. As the plots indicate, the difference between the local gradient and the true gradient increases as $H(\mathcal{D}^{(k)})$ decreases, implying that the local gradient computed by a client with more balanced data is closer to the true gradient.
\begin{figure*}[h]
    \centering
	  \subfloat[FMNIST]{
       \includegraphics[width=0.46\linewidth]{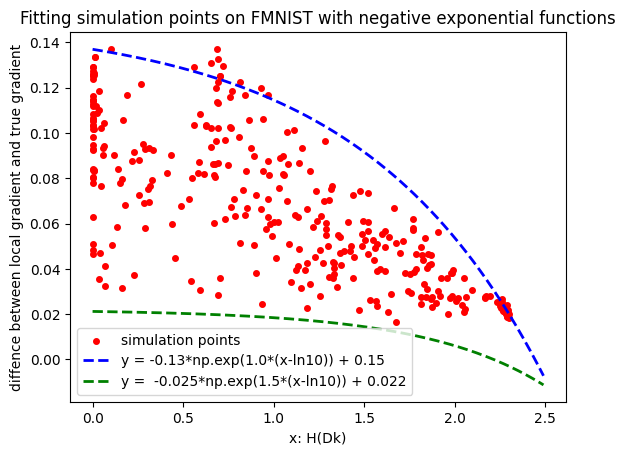}}
	  \subfloat[CIFAR10]{
        \includegraphics[width=0.46\linewidth]{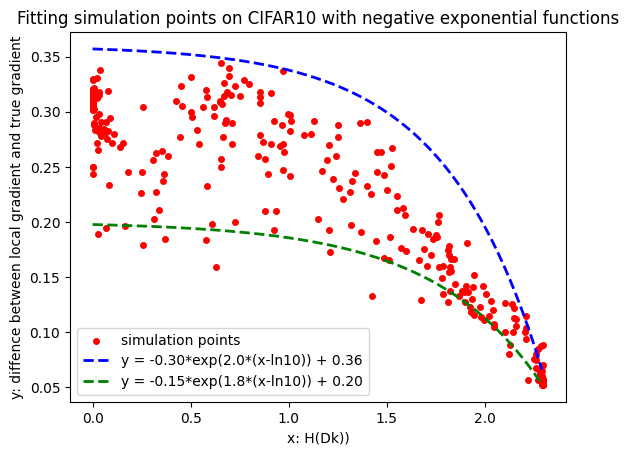}}
	  \caption{Visualization of the difference between local gradients and the global gradient (evaluated if all the data is centrally collected).}
    \label{ASSUMPTION}
\end{figure*}

\subsection{Gradient of the output (fully connected) layer's bias}
\label{gradientoffc}
Given a batch of samples $(\mathbf{x}^{(j,n)}, y^{(j,n)})$, the cross-entropy loss is readily computed as 
\begin{equation}
\label{LCE}
    \mathcal{L}_{\textbf{CE}} = -\frac{1}{Bl}\sum_{j=1}^{l}\sum_{n = 1}^{B}\log\frac{\exp\left(q^{(j,n)}_{y^{(j,n)}}\right)}{\sum_{c = 1}^{C}\exp\left(q^{(j,n)}_{c}\right)} = \frac{1}{Bl}\sum_{j = 1}^{l}\sum_{n = 1}^{B} \mathcal{L}_{\textbf{CE}}^{(j,n)}, \;\;\; y^{(j,n)} \in [C]
\end{equation}
\begin{equation}
    q^{(j,n)}_{c} = \sum_{d = 1}^{L}w_{d, c}z_{d}^{(j,n)} + b_{c},
\end{equation}
where $B$ is the batchsize; $l$ is the number of mini-batches; $C$ is the number of classes; $d$ is the dimension of the hidden space; $z_{d}^{(j,n)}$ denotes the $d$-th feature in the hidden space given sample $\mathbf{x}^{(j,n)}$ in the $j$-th batch; $w_{d,c}$ and $b_{c}$ denote the weight of $z_{d}^{(j,n)}$ and the bias for the neuron that outputs the probability of the class $c$, respectively; and $q^{(j,n)}_{c}$ is the corresponding output logit on class $c$. The gradient of the bias $b_{i}$ given sample $\mathbf{x}^{(j,n)}$ can be computed by the chain rule as
\begin{equation}
\label{grad}
    \frac{\partial \mathcal{L}_{\textbf{CE}}^{(j,n)}}{\partial b_{i}} = -\frac{\partial \mathcal{L}_{\textbf{CE}}^{(j,n)}}{\partial Q} \cdot \frac{\partial Q}{\partial q^{(j,n)}_{i}} \cdot \frac{\partial q^{(j,n)}_{i}}{\partial b_{i}},
\end{equation}
where 
\begin{equation}
    Q = \frac{\exp\left(q^{(j,n)}_{y^{(j,n)}}\right)}{\sum_{c = 1}^{C}\exp\left(q^{(j,n)}_{c}\right)}.
\end{equation}
Then
\begin{equation}
    \frac{\partial \mathcal{L}_{\textbf{CE}}^{(j,n)}}{\partial Q} = \frac{1}{Q}, \quad \frac{\partial q^{(j,n)}_{i}}{\partial b_{i}} = 1.
\end{equation}
If $i = y^{(j,n)}$,
\begin{equation}
\label{grad_true}
    \frac{\partial Q}{\partial q^{(j,n)}_{i}} = \frac{\exp\left(q^{(j,n)}_{y^{(j,n)}}\right)\left(\sum_{c = 1}^{C}\exp\left(q^{(j,n)}_{c}\right)\right) - \exp\left(q^{(j,n)}_{y^{(j,n)}}\right)^{2}}{\left(\sum_{c = 1}^{C}\exp\left(q^{(j,n)}_{c}\right)\right)^{2}} = \frac{Q\sum_{c \not = y^{(j,n)}}\exp\left(q^{(j,n)}_{c}\right)}{\sum_{c = 1}^{C}\exp\left(q^{(j,n)}_{c}\right)}.
\end{equation}
If $i \not = y^{(j,n)}$,
\begin{equation}
\label{grad_false}
    \frac{\partial Q}{\partial q^{(j,n)}_{i}} = - \frac{  \exp\left(q^{(j,n)}_{y^{(j,n)}}\right) \exp\left(q^{(j,n)}_{i}\right)}{\left(\sum_{c = 1}^{C}\exp\left(q^{(j,n)}_{c}\right)\right)^{2}} = -\frac{Q\exp\left(q^{(j,n)}_{i}\right)}{\sum_{c = 1}^{C}\exp\left(q^{(j,n)}_{c}\right)}.
\end{equation}
By plugging Eq. \ref{grad_true} and \ref{grad_false} in Eq. \ref{grad}, we obtain
\begin{equation}
\label{grad_yn}
    \frac{\partial \mathcal{L}_{\textbf{CE}}^{(j,n)}}{\partial b_{i}} =  -\frac{\sum_{c \not = y^{(j,n)}}\exp\left(q^{(j,n)}_{c}\right)}{\sum_{c = 1}^{C}\exp\left(q^{(j,n)}_{c}\right)}, \text{ if } i = y^{(j,n)};  \frac{\partial \mathcal{L}_{\textbf{CE}}^{(j,n)}}{\partial b_{i}} =  \frac{\exp\left(q^{(j,n)}_{i}\right)}{\sum_{c = 1}^{C}\exp\left(q^{(j,n)}_{c}\right)}, \text{ if } i \not = y^{(j,n)}.
\end{equation}

\subsection{Expectation of the local update $\Delta \mathbf{b}^{(k)}$}
\label{AnalysisOfexpection}
By combining Eq. \ref{grad_yn_paper} and \ref{expectationofupdate} and taking expectation, we obtain 
\begin{equation}
    \begin{aligned}
\mathbb{E}\left[\Delta b^{(k)}_{i}\right] &= -\frac{\eta}{Bl}\sum^{l}_{j=1}\sum_{n=1}^{B}\sum_{r=1}^{R}  \mathbb{E}\left[\nabla_{b_{i}} \mathcal{L}^{(j,n,r)}_{\textbf{CE}}\right]\\
&= \eta \sum_{r=1}^{R}\mathbb{P}\{i = y^{(j,n)}\}\mathbb{I}\{i = y^{(j,n)}\}\frac{1}{Bl}\sum^{l}_{j=1}\sum_{n=1}^{B}\frac{\sum_{c \not = i}\exp({q^{(j,n,r)}_{c}})}{\sum_{c = 1}^{C}\exp({q^{(j,n,r)}_{c}})} \\
& \quad -\eta\sum_{r=1}^{R}\mathbb{P}\{i \not = y^{(j,n)}\} \mathbb{I}\{i \not = y^{(j,n)}\} \frac{1}{Bl}\sum^{l}_{j=1}\sum_{n=1}^{B}\frac{\exp({q^{(j,n,r)}_{i}})}{\sum_{c = 1}^{C}\exp({q^{(j,n,r)}_{c}})}\\
&=  \eta\sum_{r=1}^{R}D^{(k)}_{i}\sum_{c \not = i}\mathbb{E}_{(\mathbf{x},y) \sim \mathcal{B}^{-c}}\left[\frac{\exp({q^{(j,n,r)}_{c}})}{\sum_{c = 1}^{C}\exp({q^{(j,n,r)}_{c}})}\right] \\
&\quad - \eta\sum_{r=1}^{R} (1-D^{(k)}_{i})\mathbb{E}_{(\mathbf{x},y) \sim \mathcal{B}^{-i}} \left[\frac{\exp({q^{(j,n,r)}_{i}})}{\sum_{c = 1}^{C}\exp({q^{(j,n,r)}_{c}})}\right]\\
&= \eta \sum_{r=1}^{R}D^{(k)}_{i}\sum_{c \not = i}\mathbb{E}_{(\mathbf{x},y ) \sim \mathcal{B}^{-c}}\left[\mathbf{s}^{-c}_{c}(\mathbf{x})\right] -\eta\sum_{r=1}^{R} (1-D^{(k)}_{i})\mathbb{E}_{(\mathbf{x},y ) \sim \mathcal{B}^{-i}} \left[\mathbf{s}^{-i}_{i}(\mathbf{x})\right]\\
&= \eta R\left(D_{i}^{(k)} \sum_{c \not = i}\mathcal{E}_{c} - (1-D^{(k)}_{i}) \mathcal{E}_{i}\right)\\
&= \eta R\left(D_{i}^{(k)} \sum_{c = 1}^{C}\mathcal{E}_{c} - \mathcal{E}_{i}\right).
    \end{aligned}
\end{equation}

Note that
\begin{equation}
    \begin{aligned}
\sum_{i=1}^{C}\mathcal{E}_{i} &= \sum_{i=1}^{C} \mathbb{E}_{(\mathbf{x},y) \sim \mathcal{B}^{-i}} \left[\mathbf{s}^{-i}_{i}(\mathbf{x})\right]\\
&= \mathbb{E}\left[\sum_{i=1}^{C} \frac{1}{C-1}\sum_{c \not = i}\frac{1}{BlD_{c}^{(k)}}\sum^{l}_{j=1}\sum_{n=1}^{B}\mathbb{I}\{ y^{(j,n)} = c\}\frac{\exp({q^{(j,n)}_{i}})}{\sum_{c = 1}^{C}\exp({q^{(j,n)}_{c}})}\right]\\
&=  \frac{1}{C-1} \sum_{i=1}^{C} \frac{1}{BlD_{i}^{(k)}}\sum^{l}_{j=1}\sum_{n=1}^{B}\mathbb{P}\{ y^{(j,n)} = i\}\frac{\sum_{c \not = i}\exp({q^{(j,n)}_{c}})}{\sum_{c = 1}^{C}\exp({q^{(j,n)}_{c}})}\\
&=    - \frac{C}{C-1}\frac{1}{Bl}\sum^{l}_{j=1}\sum_{n=1}^{B} \frac{\exp({q^{(j,n)}_{y^{(j,n)}}})}{\sum_{c = 1}^{C}\exp({q^{(j,n)}_{c}})} + \frac{C}{C-1}\\
    \end{aligned}
\end{equation}
A comparison to $\mathcal{L}_{\textbf{CE}}$ in Eq. \ref{LCE} reveals that as $\mathcal{L}_{\textbf{CE}}$ decreases during training, so does
$\sum_{i=1}^{C}\mathcal{E}_{i}$. Given an untrained/initialized neural network model,  $\mathcal{E}_{i}^{0} = 1/C\text{ for }\forall i \in [C]$, i.e., $\sum_{i=1}^{C}\mathcal{E}_{i}^{0} = -\frac{1}{C-1} + \frac{C}{C-1} = 1$. At global round $T$, if $\mathcal{L}_{\textbf{CE}}^{*} = 0$, then $\sum_{i=1}^{C}\mathcal{E}_{i}^{T} =  -\frac{C}{C-1} + \frac{C}{C-1}  = 0$.

\subsection{Privacy of $\mathcal{D}^{(k)}$}
\label{privacy}
According to Eq. \ref{expactation_b}, the server is able to obtain $C$ linear equations from each client, 
\begin{equation}
\mathbb{E}\left[\Delta b^{(k)}_{i}\right] =  \eta R \left(D_{i}^{(k)}\sum_{c=1}^{C}\mathcal{E}_{c} -\mathcal{E}_{i}\right),\text{for } \forall i \in [C], 
\end{equation}
\begin{equation}
    \sum_{i=1}^{C} D_{i}^{(k)} = 1,
\end{equation}
where $C$ denotes the number of classes. Suppose $\mathbb{E}[\Delta b^{(k)}_{i}]$ are known by the server. Then $D^{(k)}_{i}$, the variables in the aforementioned equations, cannot be determined uniquely since there are $C$ variables and $C+1$ equations. Therefore, the server is unable to infer clients' true data label distribution and the privacy of $\mathcal{D}^{(k)}$ is protected.

\subsection{Proof of Theorem \ref{theorem_estimation}}
\label{proof_theorem_1}

In Section \ref{gradientoffc} we derived an expression for the gradient of the bias in the output layer given a single sample $(\mathbf{x}^{(j,n)}, y)$ in the mini-batch. It is worthwhile making the following two observations: 
\begin{itemize}
\item the sign of the $y^{(j,n)}$-th component of
$\nabla_{\mathbf{b}} \mathcal{L}^{(j,n)}_{\textbf{CE}}(\mathbf{x}^{(j,n)}, y^{(j,n)})$ is opposite of the sign of the other components; and
\item the $y^{(j,n)}$-th component of $\nabla_{\mathbf{b}} \mathcal{L}^{(j,n)}_{\textbf{CE}}(\mathbf{x}^{(j,n)}, y^{(j,n)})$ is equal in magnitude to all other components combined.
\end{itemize}

\emph{Proof:}
Let $\Delta \mathbf{b}^{(k)} = [\Delta b^{(k)}_{1}, \dots, \Delta b^{(k)}_{C}]$ denote the local update (made by client $k$) of the bias in the output layer of the neural network model, and let $\mathcal{D}^{(k)} = [D^{(k)}_{1}, \dots, D^{(k)}_{C}]$ be the (unknown) true data label distribution, $\sum_{i=1}^{C} D^{(k)}_{i} = 1$. Assuming the learning rate $\eta$ and $R$ local epochs, the expectation of the local update of $\Delta \mathbf{b}^{(k)}$ is
\begin{equation}
    \mathbb{E}\left[\Delta b^{(k)}_{i}\right]  =  \eta R   \left(D_{i}^{(k)}\sum_{c=1}^{C}\mathcal{E}_{c} -\mathcal{E}_{i}\right).
\end{equation}
Data heterogeneity can be captured via entropy, $H(\mathcal{D}^{(k)}) = -\sum_{c = 1}^{C} D^{(k)}_{i}\ln D^{(k)}_{i}$, where higher $H(\mathcal{D}^{(k)})$ indicates that client $k$ has more balanced data. However, since we do not have access to the client's data distribution, we instead define and use as a measure of heterogeneity 
$\hat{H}(\mathcal{D}^{(k)}) \triangleq H(\text{softmax}(\Delta\mathbf{b}^{(k)}, T))$, where
\begin{equation}
     \text{softmax}(\Delta\mathbf{b}^{(k)},T)_{i} =  \frac{\exp(\Delta b_{i}^{(k)}/T)}{\sum_{c=1}^{C} \exp(\Delta b_{c}^{(k)}/T)},
\end{equation}
and where $T$ denotes the \emph{temperature} of the softmax operator. Suppose there are two clients, $u$ and $k$, with class-balanced and class-imbalanced data; let $\mathcal{D}^{(u)}$ and $\mathcal{D}^{(k)}$ denote their data label distributions, respectively, while $\hat{\mathcal{D}}^{(u)} $ and $\hat{\mathcal{D}}^{(k)}$ are computed by $\text{softmax}(\Delta\mathbf{b}^{(u)},T)$ and $\text{softmax}(\Delta\mathbf{b}^{(k)},T)$. Without a loss of generality, we can re-parameterize  $ \hat{\mathcal{D}}^{(u)}$ as 
\begin{equation}
    \hat{\mathcal{D}}^{(u)} = \epsilon\mathbf{U} + \sum_{i=1}^{C}\epsilon_{i}\mathbf{Z}_{i},
\end{equation} 
where $\mathbf{U} = [\frac{1}{C}, \dots, \frac{1}{C}]$ denotes uniform distribution; $i$-th component of $\mathbf{Z}_{i}$ is $1$ while the remaining components are $0$; $\epsilon$ and $\epsilon_{i}$ are all non-negative such that $\epsilon + \sum_{i=1}^{C}\epsilon_{i} = 1$. We can always set $\min_{j} \epsilon_{j} = 0$; otherwise, let $\epsilon^{'} = \epsilon + \min_{j} \epsilon_{j}$ and $\epsilon^{'}_{i} = \epsilon_{i} - \min_{j} \epsilon_{j}, \; \forall i \in [C]$; $\epsilon$ quantifies how close is $\hat{\mathcal{D}}^{(u)}$ to $\mathbf{U}$. Due to the concavity of entropy, 

\begin{equation}
\label{ineq1}
     H(\hat{\mathcal{D}}^{(u)}) \geq \epsilon H(\mathbf{U}) + \sum_{i=1}^{C}\epsilon_{i}H(\mathbf{Z}_{i}) = \epsilon\ln C.
\end{equation}
We will find the following lemma useful.
\begin{lemma}
\label{lemma1}
For two probability vectors \textbf{p} and \textbf{q} with dimension C, the Kullback–Leibler divergence between \textbf{p} and \textbf{q} satisfies
    \begin{equation}
        \text{KLD}(\textbf{p} || \textbf{q}) \geq \frac{1}{2} \left \Vert \mathbf{p} - \mathbf{q} \right \Vert_{1}^{2},
    \end{equation}
    where $ \left \Vert \mathbf{p} - \mathbf{q} \right \Vert_{1} = \sum_{i=1}^{C} |p_{i} - q_{i}|$.
\end{lemma}
For the proof of the lemma, please see \citep{dragomir2000some}.
Applying it, we obtain 
\begin{equation}
\label{ineq2}
    \textbf{KLD}(\hat{\mathcal{D}}^{(k)} || \mathbf{U} ) = H(\mathbf{U}) - H(\hat{\mathcal{D}}^{(k)})  \geq \frac{1}{2} \left \Vert \hat{\mathcal{D}}^{(k)} - \mathbf{U} \right \Vert_{1}^{2} \geq \frac{1}{2} \left \Vert \hat{\mathcal{D}}^{(k)} - \mathbf{U} \right \Vert_{2}^{2}.
\end{equation}
Combining Eq. \ref{ineq1} and Eq. \ref{ineq2}, we obtain
\begin{equation}
H(\hat{\mathcal{D}}^{(u)}) - H(\hat{\mathcal{D}}^{(k)}) \geq (\epsilon - 1)\ln C + \frac{1}{2} \left \Vert \hat{\mathcal{D}}^{(k)} - \mathbf{U} \right \Vert_{2}^{2}.
\end{equation}
By taking expectations of both sides,
\begin{equation}
\label{ineq4}
\mathbb{E}\left[ H(\hat{\mathcal{D}}^{(u)}) - H(\hat{\mathcal{D}}^{(k)}) \right] \geq (\mathbb{E}[\epsilon] - 1)\ln C + \frac{1}{2} \mathbb{E}\left[ \left \Vert \hat{\mathcal{D}}^{(k)} - \mathbf{U} \right \Vert_{2}^{2}\right] .
\end{equation}
Since $\left \Vert \hat{\mathcal{D}}^{(k)} - \mathbf{U} \right \Vert_{2}^{2}$ is convex (composition of the Euclidean norm and softmax), according to Jensen's inequality
\begin{equation}
\label{ineq3}
\mathbb{E}\left[ H(\hat{\mathcal{D}}^{(u)}) - H(\hat{\mathcal{D}}^{(k)}) \right] \geq (\mathbb{E}[\epsilon] - 1)\ln C + \frac{1}{2} \left \Vert \hat{\mathcal{D}}^{(k)}(\mathbb{E}[\Delta \mathbf{b}^{(k)}]) - \mathbf{U} \right \Vert_{2}^{2},
\end{equation}
where 
\begin{equation}
\hat{\mathcal{D}}^{(k)}(\mathbb{E}[\Delta \mathbf{b}^{(k)}])_{i}  = \frac{\exp\left( \eta R    \left(D^{(k)}_{i}\sum_{c=1}^{C} \mathcal{E}_{c} - \mathcal{E}_{i}\right)/T \right)}{\sum_{j}^{C}\exp\left( \eta R      \left(D^{(k)}_{j}\sum_{c=1}^{C} \mathcal{E}_{c} - \mathcal{E}_{j}\right)/T \right)}.
\end{equation}
Selecting $T$ such that $\eta R    \left(D^{(k)}_{i}\sum_{c=1}^{C} \mathcal{E}_{c} - \mathcal{E}_{i}\right)/T $ is sufficiently small and applying the first-order Taylor's expansion of $e^{x}$ around $0$, we obtain
\begin{equation}
\sum_{j}^{C}\exp\left( \eta R     \left(D^{(k)}_{j}\sum_{c=1}^{C} \mathcal{E}_{c} - \mathcal{E}_{j}\right)/T \right) =  \sum_{j}^{C} 1 +  \eta R     \sum_{j}^{C} \left(D^{(k)}_{j}\sum_{c=1}^{C} \mathcal{E}_{c} - \mathcal{E}_{j}\right)/T = C,
\end{equation}
where $\sum_{j=1}^{C} D^{(k)}_{j} = 1$. This leads to a simplified $\hat{\mathcal{D}}^{(k)}(\mathbb{E}[\Delta \mathbf{b}^{(k)}])$,
\begin{equation}
\label{simply}
\hat{\mathcal{D}}^{(k)}(\mathbb{E}[\Delta \mathbf{b}^{(k)}])_{i} = \frac{1 +  \eta R      \left(D^{(k)}_{i}\sum_{c=1}^{C} \mathcal{E}_{c} - \mathcal{E}_{i}\right)/T}{C}.
\end{equation}
Substituting Eq. \ref{simply} for the second term on the right-hand side of ineq. \ref{ineq3} leads to
\begin{equation}
     \left \Vert \hat{\mathcal{D}}^{(k)}(\mathbb{E}[\Delta \mathbf{b}^{(k)}]) - \mathbf{U} \right \Vert_{2}^{2} = \left(\frac{ \eta R     }{CT}\right)^{2}\sum_{i=1}^{C} \left( D^{(k)}_{i}\sum_{c=1}^{C} \mathcal{E}_{c} - \mathcal{E}_{i} \right)^{2}.
\end{equation}
Now, consider
\begin{equation}
     \hat{\mathcal{D}}^{(u)} - \mathbf{U} = (\epsilon - 1)\mathbf{U} + \sum_{i=1}^{C} \epsilon_{i}\mathbf{Z}_{i}.
\end{equation}
Taking expectations of both sides,
\begin{equation}
     \mathbb{E} \left[ (\epsilon - 1)\mathbf{U} + \sum_{i=1}^{C} \epsilon_{i}\mathbf{Z}_{i}\right]=  \mathbb{E} \left[\hat{\mathcal{D}}^{(u)} - \mathbf{U} \right] \geq \hat{\mathcal{D}}^{(u)}(\mathbb{E}[\Delta \mathbf{b}^{(u)}]) - \mathbf{U}.
\end{equation}
The above inequality holds component-wise, so for the $j$-component ($\epsilon_{j} = 0$)
\begin{equation}
    \mathbb{E}[\frac{1}{C}(\epsilon - 1) + \epsilon_{j}] = \mathbb{E}[\frac{1}{C}(\epsilon - 1) ] \geq   \hat{\mathcal{D}}^{(u)}(\mathbb{E}[\Delta \mathbf{b}^{(u)}])_{j}  - \mathbf{U}_{i} = \frac{ \eta R      \left(D^{(u)}_{j}\sum_{c=1}^{C} \mathcal{E}_{c} - \mathcal{E}_{j}\right)}{CT}.
\end{equation}
Therefore,
\begin{equation}
     \mathbb{E}[\epsilon]  - 1 \geq  \frac{ \eta R     \left(D^{(u)}_{j}\sum_{c=1}^{C} \mathcal{E}_{c} - \mathcal{E}_{j}\right)}{T} \geq  \min_{i} \frac{ \eta R     \left(D^{(u)}_{i}\sum_{c=1}^{C} \mathcal{E}_{c} - \mathcal{E}_{i}\right)}{T}.
\end{equation}
Taking absolute value of both sides yields
\begin{equation}
    |\mathbb{E}[\epsilon]  - 1| \leq \frac{ \eta R    }{T} \max_{i} \left| D^{(u)}_{i}\sum_{c=1}^{C} \mathcal{E}_{c} - \mathcal{E}_{i} \right| = \frac{ \eta R     }{T} \max_{i} \left| (D^{(u)}_{i} - \frac{1}{C})\sum_{c=1}^{C} \mathcal{E}_{c} - \mathcal{E}_{i} + \frac{1}{C}\sum_{c=1}^{C} \mathcal{E}_{c}\right|.
\end{equation}
By applying the triangle inequality we obtain
\begin{equation}
     |\mathbb{E}[\epsilon]  - 1| \leq \frac{ \eta R     }{T} \max_{i} \left|  D^{(u)}_{i} - \frac{1}{C}\right|\sum_{c=1}^{C} \mathcal{E}_{c} +  \frac{ \eta R     }{T}  \max_{i} \left|  \frac{1}{C}\sum_{c=1}^{C} \mathcal{E}_{c} - \mathcal{E}_{i} \right|.
\end{equation}
Let $\delta = \max_{i} \left|  \frac{1}{C}\sum_{c=1}^{C} \mathcal{E}_{c} - \mathcal{E}_{i} \right|$. Since $\sum_{c=1}^{C}\mathcal{E}_{c} \leq C\frac{1}{C} = 1$, it holds that
\begin{equation}
    |\mathbb{E}[\epsilon]  - 1| \leq \frac{ \eta R     }{T} \max_{i} \left|  D^{(u)}_{i} - \frac{1}{C}\right| + \frac{ \eta R     }{T} \delta.
\end{equation}
Furthermore, since $\mathbb{E}[\epsilon] - 1 < 0$,
\begin{equation}
      \mathbb{E}[\epsilon] - 1 \geq -\frac{ \eta R    }{T} \max_{i} \left|  D^{(u)}_{i} - \frac{1}{C}\right| - \frac{ \eta R     }{T} \delta.
\end{equation}
Note that
\begin{equation}
\begin{aligned}
    \left( D^{(k)}_{i}\sum_{c=1}^{C} \mathcal{E}_{c} - \mathcal{E}_{i} \right)^{2} &= \left((D^{(k)}_{i} - \frac{1}{C})\sum_{c=1}^{C} \mathcal{E}_{c} - \mathcal{E}_{i} + \frac{1}{C}\sum_{c=1}^{C} \mathcal{E}_{c} \right)^{2}\\
    &=\left((D^{(k)}_{i} - \frac{1}{C})\sum_{c=1}^{C} \mathcal{E}_{c} \right)^{2} + \left(\frac{1}{C}\sum_{c=1}^{C} \mathcal{E}_{c} - \mathcal{E}_{i} \right)^{2} \\
    &\quad + 2\left(\sum_{c=1}^{C}\mathcal{E}_{c}\right)\left(D^{(k)}_{i} - \frac{1}{C}\right)\left(\frac{1}{C}\sum_{c=1}^{C} \mathcal{E}_{c} - \mathcal{E}_{i}\right)\\
    &\geq \left((D^{(k)}_{i} - \frac{1}{C})\sum_{c=1}^{C} \mathcal{E}_{c} \right)^{2} + 2\left(\sum_{c=1}^{C}\mathcal{E}_{c}\right)\left(D^{(k)}_{i} - \frac{1}{C}\right)\left(\frac{1}{C}\sum_{c=1}^{C} \mathcal{E}_{c} - \mathcal{E}_{i}\right).
\end{aligned}
\end{equation}
Therefore,
\begin{equation}
\begin{aligned}
    \sum_{i=1}^{C} \left( D^{(k)}_{i}\sum_{c=1}^{C} \mathcal{E}_{c} - \mathcal{E}_{i} \right)^{2} &\geq \left(\sum_{c=1}^{C} \mathcal{E}_{c} \right)^{2} \sum_{i=1}^{C} \left(D^{(k)}_{i} - \frac{1}{C}\right)^{2}\\
    & \quad + 2\left(\sum_{c=1}^{C}\mathcal{E}_{c}\right)\sum_{i=1}^{C}\left(D^{(k)}_{i} - \frac{1}{C}\right)\left(\frac{1}{C}\sum_{c=1}^{C} \mathcal{E}_{c} - \mathcal{E}_{i}\right)\\
    &= \left(\sum_{c=1}^{C} \mathcal{E}_{c} \right)^{2} \sum_{i=1}^{C} \left(D^{(k)}_{i} - \frac{1}{C}\right)^{2}\\
    &\quad + 2\left(\sum_{c=1}^{C}\mathcal{E}_{c}\right)\sum_{i=1}^{C}\left(\frac{D^{(k)}_{i}}{C}\sum_{c=1}^{C} \mathcal{E}_{c}  - \frac{1}{C^{2}}\sum_{c=1}^{C} \mathcal{E}_{c} + \frac{\mathcal{E}_{i}}{C}\ - D_{i}^{(k)}\mathcal{E}_{i} \right)\\
    &= \left(\sum_{c=1}^{C} \mathcal{E}_{c} \right)^{2} \sum_{i=1}^{C} \left(D^{(k)}_{i} - \frac{1}{C}\right)^{2}\\
    &\quad + 2\left(\sum_{c=1}^{C}\mathcal{E}_{c}\right)\left(\frac{1}{C}\sum_{c=1}^{C} \mathcal{E}_{c}  - \frac{1}{C}\sum_{c=1}^{C} \mathcal{E}_{c} + \frac{1}{C}\sum_{i=1}^{C} \mathcal{E}_{i} - \sum_{i=1}^{C}D_{i}^{(k)}\mathcal{E}_{i} \right)\\
    &\geq \left(\sum_{c=1}^{C} \mathcal{E}_{c} \right)^{2} \sum_{i=1}^{C} \left(D^{(k)}_{i} - \frac{1}{C}\right)^{2} + 2\left(\sum_{c=1}^{C}\mathcal{E}_{c}\right)\left( \frac{1}{C}\sum_{c=1}^{C} \mathcal{E}_{c} - \max_{j}\mathcal{E}_{j} \right)\\
    &\geq \left(\sum_{c=1}^{C} \mathcal{E}_{c} \right)^{2} \sum_{i=1}^{C} \left(D^{(k)}_{i} - \frac{1}{C}\right)^{2} - 2\delta.
\end{aligned}
\end{equation}
Substituting the above expression in Eq. \ref{ineq3}, we obtain
\begin{align}
\label{ineq5}
    \mathbb{E}\left[ H(\hat{\mathcal{D}}^{(u)}) - H(\hat{\mathcal{D}}^{(k)}) \right] &\geq -\frac{ \eta R \ln C   }{T} \max_{j} \left|  D^{(u)}_{j} - \frac{1}{C}\right| - \frac{ \eta R  \ln C }{T} \delta \\
    &+ \frac{1}{2}  \left(\frac{ \eta R     }{CT}\right)^{2} \left(\sum_{c=1}^{C} \mathcal{E}_{c} \right)^{2} \sum_{i=1}^{C} \left(D^{(k)}_{i} - \frac{1}{C}\right)^{2} -  \left(\frac{ \eta R}{CT}\right)^{2}\delta,
\end{align}
and, therefore,
\begin{align}
\label{ineq6}
    \mathbb{E}\left[ H(\hat{\mathcal{D}}^{(u)}) - H(\hat{\mathcal{D}}^{(k)}) \right] &\geq 
    \frac{1}{2}  \left(\frac{ \eta R     }{CT}\sum_{c=1}^{C} \mathcal{E}_{c}\right)^{2} \left \Vert \mathcal{D}^{(k)} - \mathbf{U}\right\Vert_{2}^{2} -\frac{ \eta R \ln C  }{T} \left\Vert  \mathcal{D}^{(u)}  - \mathbf{U}\right\Vert_{\infty} -\mathcal{C}\delta, 
\end{align}
where $\mathcal{C} = \frac{ \eta R (\eta R + C^{2}T\ln C) }{C^{2}T^{2}}$.
$\hfill \blacksquare$

\subsection{Convergence Analysis}
\label{converge}
Here we present the convergence analysis of an FL system deploying FedAvg with SGD wherein only a small fraction of clients participates in any given round of training. Recall that the objective function that comes up when training a neural network model is generally non-convex; we make the standard assumptions of smoothness, unbiased gradient estimate, and bounded variance.

\begin{assumption}[Smoothness]
\label{smooth}
    Each local objective function $F_{k}(\cdot)$ is $L$-smooth,
    \begin{equation}
        \left\Vert \nabla F_{k}(\theta^{t+1}_{k}) -  \nabla F_{k}(\theta^{t}_{k})\right\Vert_{2} \leq L  \left\Vert \theta^{t+1}_{k} - \theta^{t}_{k}\right\Vert_{2}.
    \end{equation}
\end{assumption}

\begin{assumption}[Gradient oracle]
\label{gradient}
The stochastic gradient estimator $g_{k}(\theta^{t, r}_{k}) = \nabla F_{k}(\theta^{t,r}_{k}) + \zeta^{t,r}_{k}$ for each global round $t$ and local epoch $r$ is such that 
\begin{equation}
     \mathbb{E}[\zeta^{t,r}_{k}] = 0
\end{equation}
and 
\begin{equation}
     \mathbb{E}\left[\left\Vert \zeta^{t,r}_{k} \right\Vert^{2} | \theta_{k}^{t,r}\right] \leq  \sigma^{2}.
\end{equation}
\end{assumption}

With these three assumptions in place, we provide the proof of 
Theorem \ref{theorem 2} stated in the main paper. The proof relies on the
technique previously used in \citep{chen2020optimal, yang2021achieving}, where the sampling 
method is unbiased and thus $\mathbb{E}\left[\frac{1}{K}\sum_{k \in \mathcal{S}^{t}}\sum_{r=1}^{R} g_{k}(\theta_{k}^{t,r})\right] = \sum_{k=1}^{N}\sum_{r=1}^{R}p_{k} \nabla F_{k}(\theta_{k}^{t,r})$. 
We provide a generalization that holds for any sampling strategy, resulting in $\mathbb{E}\left[\frac{1}{K}\sum_{k \in \mathcal{S}^{t}}\sum_{r=1}^{R}g_{k}(\theta_{k}^{t,r})\right] = \sum_{k=1}^{N}\sum_{r=1}^{R}\omega_{k}^{t}\nabla F_{k}(\theta_{k}^{t,r})$, where $\omega_{k}^{t}$ denotes the probability 
of sampling client $k$ in round $t$ under sampling strategy $\Pi$. Note that $\sum_{k=1}^{N} \omega_{k}^{t} = 1$.
We assume that all clients deploy the same number of local epochs $R$ and use learning rate $\eta$ at round $t$.

\subsubsection{key lemma}
\begin{lemma}
\label{lemma2}
(Lemma 2 in \citep{yang2021achieving}) Instate Assumptions \ref{bounded_disimilarity}, \ref{smooth} and \ref{gradient}. For any step size $\eta$ such that $\eta \leq \frac{1}{8LR}$, for any client $k$ it holds that
\begin{equation}
    \mathbb{E}\left[ \left \Vert \theta_{k}^{t,r} - \theta^{t}\right \Vert^{2} \right] \leq 5R\eta^{2}(\sigma^{2} + 6R\sigma_{k}^{2}) + 30R^{2}\eta^{2}\left\Vert\nabla F(\theta^{t})\right\Vert^{2}.
\end{equation}
\end{lemma}
\emph{Proof of Lemma \ref{lemma2}:}
For any client $k \in [N]$ and $r \in [R]$,
\begin{equation}
    \begin{aligned}
 \mathbb{E}\left[ \left \Vert \theta_{k}^{t,r} - \theta^{t}\right \Vert^{2} \right] &= \mathbb{E}\left[ \left \Vert \theta_{k}^{t,r-1} - \theta^{t}  -\eta g_{k}( \theta_{k}^{t,r-1})\right \Vert^{2}\right] \\
 &= \mathbb{E}[  \Vert \theta_{k}^{t,r-1} - \theta^{t}  -\eta ( g_{k}( \theta_{k}^{t,r-1}) - \nabla F_{k}(\theta_{k}^{t,r-1}) + \nabla F_{k}(\theta_{k}^{t,r-1}) \\
 & \quad - \nabla F_{k}(\theta^{t}) + \nabla F_{k}(\theta^{t}) - \nabla F(\theta^{t}) + \nabla F(\theta^{t}) ) \Vert^{2}] \\
 &\leq\left(1+ \frac{1}{2R-1}\right)\mathbb{E} \left\Vert \theta_{k}^{t,r-1} - \theta^{t}\right \Vert^{2} + \eta^{2}\mathbb{E} \left\Vert g_{k}(\theta_{k}^{t,r-1}) - \nabla F_{k}(\theta_{k}^{t,r-1})\right \Vert^{2}\\
 &\quad + 6R\eta^{2}\mathbb{E} \left\Vert \nabla F_{k}(\theta_{k}^{t,r-1}) -  \nabla F_{k}(\theta^{t})\right \Vert^{2} + 6R\eta^{2}\mathbb{E} \left\Vert g_{k}(  \nabla F_{k}(\theta^{t}) -  \nabla F(\theta^{t})\right \Vert^{2}\\
 &\quad +6R\eta^{2}\mathbb{E} \left\Vert \nabla F(\theta^{t})\right \Vert^{2} \\
 & \leq   \left(1+ \frac{1}{2R-1}\right)\mathbb{E} \left\Vert \theta_{k}^{t,r-1} - \theta^{t}\right \Vert^{2} + \eta^{2}\sigma^{2} + 6R\eta^{2}L^{2}\mathbb{E} \left\Vert \theta_{k}^{t,r-1} - \theta^{t}\right \Vert^{2}\\
 &\quad +  6R\eta^{2}\sigma_{k}^{2} +6R\eta^{2}\mathbb{E} \left\Vert \nabla F(\theta^{t})\right \Vert^{2} \\
 &=  \left(1+ \frac{1}{2R-1} + 6R\eta^{2}L^{2}\right)\mathbb{E} \left\Vert \theta_{k}^{t,r-1} - \theta^{t}\right \Vert^{2} + \eta^{2}\sigma^{2} + 6R\eta^{2}\sigma_{k}^{2} \\
 &\quad+6R\eta^{2}\mathbb{E} \left\Vert \nabla F(\theta^{t})\right \Vert^{2}\\
 &\leq \left(1+ \frac{1}{R-1}\right)\mathbb{E} \left\Vert \theta_{k}^{t,r-1} - \theta^{t}\right \Vert^{2} + \eta^{2}\sigma^{2} + 6R\eta^{2}\sigma_{k}^{2} +6R\eta^{2}\mathbb{E} \left\Vert \nabla F(\theta^{t})\right \Vert^{2}.
    \end{aligned}
\end{equation}

Unrolling the recursion yields
\begin{equation}
    \begin{aligned}
 \mathbb{E}\left[ \left \Vert \theta_{k}^{t,r} - \theta^{t}\right \Vert^{2} \right] &\leq \sum_{r=1}^{R}\left(1+ \frac{1}{R-1}\right)^{r-1}\left(\eta^{2}\sigma^{2} + 6R\eta^{2}\sigma_{k}^{2} +6R\eta^{2}\mathbb{E} \left\Vert \nabla F(\theta^{t})\right \Vert^{2}\right)\\
 &\leq (R-1)\left[\left(1 + \frac{1}{R-1}\right)^{R} - 1\right]\left(\eta^{2}\sigma^{2} + 6R\eta^{2}\sigma_{k}^{2} +6R\eta^{2}\mathbb{E} \left\Vert \nabla F(\theta^{t})\right \Vert^{2}\right)\\
 &\leq 5R\eta^{2}\left(\sigma^{2} + 6R\sigma_{k}^{2}\right) + 30R^{2}\eta^{2} \left\Vert \nabla F(\theta^{t})\right \Vert^{2}.
    \end{aligned}
\end{equation}
$\hfill \blacksquare$

\subsubsection{Proof of Theorem \ref{theorem 2}}
The model update at global round $t$ is formed as
\begin{equation}
    \theta^{t+1} = \theta^{t} -\eta\frac{1}{K}\sum_{k\in \mathbf{S}^{t}}\sum_{r=1}^{R} g_{k}(\theta_{k}^{t,r}), 
\end{equation}

where $\theta^{t+1}$ and $\theta^{t}$ denote parameters of 
the global model at rounds $t+1$ and $t$, respectively, and $\theta_{k}^{t,r}$ denotes parameters of the local model of 
client $k$ after $r$ local training epochs. Let 
\begin{equation}
    \Delta^{t} \triangleq  \frac{1}{K}\sum_{k\in \mathbf{S}^{t}}\sum_{r=1}^{R} g_{k}(\theta_{k}^{t,r}). 
\end{equation}
Taking the expectations (conditioned on $\theta^{t}$) of both sides, 
we obtain
\begin{equation}
\label{inequalA}
\begin{aligned}
    \mathbb{E}\left[ F(\theta^{t+1})\right] &= \mathbb{E}\left[ F(\theta^{t} -  \eta\Delta^{t} )\right]\\
    &\stackrel{(a)}{\leq} F(\theta^{t}) - \eta\left<\nabla F(\theta^{t}), \mathbb{E}\left[  \Delta^{t} \right] \right> + \frac{L}{2}\eta^{2}\mathbb{E}\left[\left\Vert\Delta^{t}\right\Vert^{2}\right]\\
    &= F(\theta^{t}) + \eta\left<\nabla F(\theta^{t}),    \mathbb{E}\left[  R\nabla F(\theta^{t}) - R\nabla F(\theta^{t}) - \Delta^{t} \right] \right> + \frac{L}{2}\eta^{2}\mathbb{E}\left[\left\Vert\Delta^{t}\right\Vert^{2}\right]\\
    &= F(\theta^{t}) - R \eta  \left\Vert\nabla F(\theta^{t})\right\Vert^{2} +\eta \underbrace{\left<\nabla F(\theta^{t}),  \mathbb{E}\left[ R\nabla F(\theta^{t}) - \Delta^{t} \right] \right>}_{A_{1}} + \frac{L}{2}\eta^{2}\underbrace{\mathbb{E}\left[\left\Vert\Delta^{t}\right\Vert^{2}\right]}_{A_{2}}.\\
\end{aligned}    
\end{equation}
Inequality (a) in the expression above holds due to the 
smoothness of $F(\cdot)$ (see Assumption \ref{smooth}). Note that the term $A_1$ can be bounded as
\begin{equation}
\begin{aligned}
A_1 &= \left<\nabla F(\theta^{t}), \mathbb{E}\left[R\nabla F(\theta^{t}) -   \Delta^{t} \right] \right>\\
&=  \left<\nabla F(\theta^{t}), \mathbb{E}\left[R\nabla F(\theta^{t}) - \frac{1}{K}\sum_{k \in \mathcal{S}^{t}}\sum_{r=1}^{R}g_{k}(\theta_{k}^{t,r})\right]\right>\\
&=  \left<\nabla F(\theta^{t}), \mathbb{E}\left[R\nabla F(\theta^{t})\right] - \sum_{k=1}^{N}\sum_{r=1}^{R}\omega_{k}^{t}\nabla F_{k}(\theta_{k}^{t,r})\right>\\
&= \sum_{k=1}^{N}\omega_{k}^{t} \left<\sqrt{R}\nabla F(\theta^{t}),  - \frac{1}{\sqrt{R}}\mathbb{E}\left[\sum_{r=1}^{R}\left(\nabla F_{k}(\theta_{k}^{t,r}) - \nabla F(\theta^{t})\right)\right] \right>\\
&\stackrel{(a)}{=} \frac{R}{2}\left\Vert \nabla F(\theta^{t})  \right\Vert^{2} + \frac{1}{2R}\sum_{k=1}^{N}\omega_{k}^{t}\mathbb{E}\left\Vert \sum_{r=1}^{R}\left(\nabla F_{k}(\theta_{k}^{t,r}) - \nabla F(\theta^{t})\right)\right\Vert^{2} - \frac{1}{2R}\sum_{k=1}^{N}\omega_{k}^{t}\mathbb{E}\left\Vert\sum_{r=1}^{R}\nabla F_{k}(\theta_{k}^{t,r})\right\Vert^{2}\\
&\stackrel{(b)}{\leq} \frac{R}{2}\left\Vert \nabla F(\theta^{t})  \right\Vert^{2} + \frac{1}{R}\sum_{k=1}^{N}\omega_{k}^{t}\mathbb{E}\left\Vert \sum_{r=1}^{R}\left(\nabla F_{k}(\theta_{k}^{t,r}) - \nabla F_{k}(\theta^{t})\right)\right\Vert^{2} 
 \\
&\quad + \frac{1}{R}\sum_{k=1}^{N}\omega_{k}^{t}\mathbb{E}\left\Vert \sum_{r=1}^{R}\left(\nabla F_{k}(\theta^{t}) - \nabla F(\theta^{t})\right)\right\Vert^{2} -  \frac{1}{2R}\sum_{k=1}^{N}\omega_{k}^{t}\mathbb{E}\left\Vert\sum_{r=1}^{R}\nabla F_{k}(\theta_{k}^{t,r})\right\Vert^{2}\\
&\stackrel{(c)}{\leq}  \frac{R}{2}\left\Vert \nabla F(\theta^{t})  \right\Vert^{2} + \sum_{k=1}^{N}\omega_{k}^{t}\sum_{r=1}^{R}\mathbb{E}\left\Vert\nabla F_{k}(\theta_{k}^{t,r}) - \nabla F_{k}(\theta^{t})\right\Vert^{2} \\
&\quad + \sum_{k=1}^{N}\omega_{k}^{t}\sum_{r=1}^{R}\mathbb{E}\left\Vert\nabla F_{k}(\theta^{t}) - \nabla F(\theta^{t})\right\Vert^{2} - \frac{1}{2R}\sum_{k=1}^{N}\omega_{k}^{t}\mathbb{E}\left\Vert\sum_{r=1}^{R}\nabla F_{k}(\theta_{k}^{t,r})\right\Vert^{2}\\
&\stackrel{(d)}{\leq} \frac{R}{2}\left\Vert \nabla F(\theta^{t})  \right\Vert^{2} + L^{2}\sum_{k=1}^{N}\omega_{k}^{t}\sum_{r=1}^{R}\mathbb{E}\left\Vert\theta_{k}^{t,r} - \theta^{t}\right\Vert^{2} + R\sum_{k=1}^{N}\omega_{k}^{t}\sigma_{k}^{2} - \frac{1}{2R}\sum_{k=1}^{N}\omega_{k}^{t}\mathbb{E}\left\Vert\sum_{r=1}^{R}\nabla F_{k}(\theta_{k}^{t,r})\right\Vert^{2},\\
\end{aligned}
\end{equation}
where equality (a) follows from $\left<\mathbf{x}, \mathbf{y}\right> = \frac{1}{2}\left(\Vert\mathbf{x}\right\Vert^{2} + \left\Vert\mathbf{y}\right\Vert^{2} - \left\Vert\mathbf{x-y}\right\Vert^{2})$, inequality (b) is due to $  \left\Vert\mathbf{x+y}\right\Vert^{2} \leq 2\left\Vert\mathbf{x}\right\Vert^{2} + 2\left\Vert\mathbf{y}\right\Vert^{2}$, inequality (c) holds because $\left\Vert \sum_{i=1}^{n}\mathbf{z}_{i}\right \Vert^{2} \leq n\sum_{i=1}^{n}\left\Vert \mathbf{z}_{i}\right \Vert^{2}$, and inequality (d) follows from Assumptions \ref{bounded_disimilarity} and \ref{smooth}. By selecting $\eta < \frac{1}{8LR}$ and applying Lemma \ref{lemma2} we obtain
\begin{equation}
\label{ineqA1}
    \begin{aligned}
A_1 & \leq \frac{R}{2}\left\Vert \nabla F(\theta^{t})  \right\Vert^{2} + L^{2}\sum_{k=1}^{N}\omega_{k}^{t}\sum_{r=1}^{R}\left[5R\eta^{2}(\sigma^{2} + 6R\sigma_{k}^{2}) + 30R^{2}\eta^{2}\left\Vert\nabla F(\theta^{t})\right\Vert^{2}\right]\\
&\quad +R\sum_{k=1}^{N}\omega_{k}^{t}\sigma_{k}^{2} - \frac{1}{2R}\sum_{k=1}^{N}\omega_{k}^{t}\mathbb{E}\left\Vert\sum_{r=1}^{R}\nabla F_{k}(\theta_{k}^{t,r})\right\Vert^{2}\\
&= \left(\frac{R}{2} + 30L^{2}R^{3}\eta^{2}\right)\left\Vert \nabla F(\theta^{t})  \right\Vert^{2} + 5L^{2}R^{2}\eta^{2}\sigma^{2} + 30L^{2}R^{3}\eta^{2}\sum_{k=1}^{N}\omega_{k}^{t}\sigma_{k}^{2}\\
&\quad +R\sum_{k=1}^{N}\omega_{k}^{t}\sigma_{k}^{2} - \frac{1}{2R}\sum_{k=1}^{N}\omega_{k}^{t}\mathbb{E}\left\Vert\sum_{r=1}^{R}\nabla F_{k}(\theta_{k}^{t,r})\right\Vert^{2}.\\
    \end{aligned}
\end{equation}
Furthermore, 
\begin{equation}
\label{ineqA2}
    \begin{aligned}
A_{2} 
&= \mathbb{E}\left[\left\Vert\frac{1}{K}\sum_{k \in \mathcal{S}^{t}}\sum_{r=1}^{R}g_{k}(\theta_{k}^{t,r})\right\Vert^{2}\right]\\
&= \mathbb{E}\left[\left\Vert\sum_{k=1}^{N}\frac{\mathbb{I}\{k \in \mathcal{S}^{t}\}}{K}\sum_{r=1}^{R}g_{k}(\theta_{k}^{t,r})\right\Vert^{2}\right]\\
&= \mathbb{E}\left[\left\Vert\sum_{k=1}^{N}\frac{\mathbb{I}\{k \in \mathcal{S}^{t}\}}{K}\sum_{r=1}^{R}g_{k}(\theta_{k}^{t,r}) - \nabla F_{k}(\theta_{k}^{t,r}) + \nabla F_{k}(\theta_{k}^{t,r})\right\Vert^{2}\right]\\
&\stackrel{(a)}{=} \mathbb{E}\left[\left\Vert\sum_{k=1}^{N}\frac{\mathbb{I}\{k \in \mathcal{S}^{t}\}}{K}\sum_{r=1}^{R}g_{k}(\theta_{k}^{t,r}) - \nabla F_{k}(\theta_{k}^{t,r})\right\Vert^{2}\right] + \mathbb{E}\left[\left\Vert\sum_{k=1}^{N}\frac{\mathbb{I}\{k \in \mathcal{S}^{t}\}}{K}\sum_{r=1}^{R}\nabla F_{k}(\theta_{k}^{t,r})\right\Vert^{2}\right]\\
&\stackrel{(b)}{\leq} \mathbb{E}\left[\sum_{k=1}^{N}\frac{\mathbb{I}\{k \in \mathcal{S}^{t}\}}{K}\sum_{r=1}^{R} \left\Vert g_{k}(\theta_{k}^{t,r}) - \nabla F_{k}(\theta_{k}^{t,r})\right\Vert^{2}\right] + \mathbb{E}\left[\sum_{k=1}^{N}\frac{\mathbb{I}\{k \in \mathcal{S}^{t}\}}{K}\left\Vert\sum_{r=1}^{R}\nabla F_{k}(\theta_{k}^{t,r})\right\Vert^{2}\right]\\
&\stackrel{(c)}{\leq} R\sigma^2 + \sum_{k=1}^{N}\omega_{k}^{t}\mathbb{E}\left\Vert\sum_{r=1}^{R}\nabla F_{k}(\theta_{k}^{t,r})\right\Vert^{2},
    \end{aligned}
\end{equation}
where equation (a) holds because $\mathbb{E}\left[ g_{k}(\theta_{k}^{t,r}) - \nabla F_{k}(\theta_{k}^{t,r})\right] = 0$, inequality (b) stems from the Jensen's inequality, and inequality (c) is due to Assumption \ref{gradient}.

Substituting inequalities (\ref{ineqA1}) and (\ref{ineqA2}) into inequality (\ref{inequalA}) yields
\begin{equation}
    \begin{aligned}
\mathbb{E}\left[ F(\theta^{t+1})\right] &\leq F(\theta^{t}) - R \eta  \left\Vert\nabla F(\theta^{t})\right\Vert^{2} +\eta \underbrace{\left<\nabla F(\theta^{t}),  \mathbb{E}\left[ R\nabla F(\theta^{t}) - \Delta^{t} \right] \right>}_{A_{1}} + \frac{L}{2}\eta^{2}\underbrace{\mathbb{E}\left[\left\Vert\Delta^{t}\right\Vert^{2}\right]}_{A_{2}}\\
&\leq  F(\theta^{t}) - R\eta\left(\frac{1}{2} - 30L^{2}R^{2}\eta^{2}\right)\left\Vert\nabla F(\theta^{t})\right\Vert^{2} + \left( 5L^{2}R^{2}\eta^{3} + \frac{LR}{2}\eta^{2}\right)\sigma^{2} \\
&+ \left(30L^{2}R^{3}\eta^{3} + R\eta\right)\sum_{k=1}^{N}\omega_{k}^{t}\sigma_{k}^{2} + \left(\frac{L}{2}\eta^{2} - \frac{\eta}{2R}\right) \sum_{k=1}^{N}\omega_{k}^{t}\mathbb{E}\left\Vert\sum_{r=1}^{R}\nabla F_{k}(\theta_{k}^{t,r})\right\Vert^{2}.
    \end{aligned}
\end{equation}
If $\eta < \frac{1}{8LR}$, it must be that $\frac{1}{2} - 30L^{2}R^{2}\eta^{2} > 0$ and $\frac{L}{2}\eta^{2} - \frac{\eta}{2R} < 0$, leading to
\begin{equation}
\begin{aligned}
\mathbb{E}\left[ F(\theta^{t+1})\right] 
&\leq  F(\theta^{t}) - R\eta\left(\frac{1}{2} - 30L^{2}R^{2}\eta^{2}\right)\left\Vert\nabla F(\theta^{t})\right\Vert^{2} \\
&+ \left( 5L^{2}R^{2}\eta^{3} + \frac{LR}{2}\eta^{2}\right)\sigma^{2} + \left(30L^{2}R^{3}\eta^{3} + R\eta\right)\sum_{k=1}^{N}\omega_{k}^{t}\sigma_{k}^{2}.
\end{aligned}
\end{equation}

By rearranging and summing from $t = 0$ to $t =  \mathcal{T}-1$ we obtain
\begin{equation}
\begin{aligned}
\mathbb{E}\left[ F(\theta^{\mathcal{T}})\right] - F(\theta^{0}) &\leq -R\eta\left(\frac{1}{2} - 30L^{2}R^{2}\eta^{2}\right)\sum_{t=0}^{\mathcal{T}-1} \left\Vert\nabla F(\theta^{t})\right\Vert^{2}\\
&+ \left( 5L^{2}R^{2}\eta^{3} + \frac{LR}{2}\eta^{2}\right)\mathcal{T}\sigma^{2} + \left(30L^{2}R^{3}\eta^{3} + R\eta\right)\sum_{t=0}^{\mathcal{T}-1}\sum_{k=1}^{N}\omega_{k}^{t}\sigma_{k}^{2} \\
&\leq -R\eta\left(\frac{1}{2} - 30L^{2}R^{2}\eta^{2}\right)\mathcal{T} \min_{t \in [\mathcal{T}]} \left\Vert\nabla F(\theta^{t})\right\Vert^{2}\\
&+ \left( 5L^{2}R^{2}\eta^{3} + \frac{LR}{2}\eta^{2}\right)\mathcal{T}\sigma^{2} + \left(30L^{2}R^{3}\eta^{3} + R\eta\right)\sum_{t=0}^{\mathcal{T}-1}\sum_{k=1}^{N}\omega_{k}^{t}\sigma_{k}^{2}. \\
\end{aligned}
\end{equation}
Let $\theta^{*}$ denote the optimal model's parameters, i.e., $F(\theta^{*}) \leq F(\theta^{t}) \forall t \in [\mathcal{T}]$. Then
\begin{equation}
\begin{aligned}
\min_{t \in [\mathcal{T}]} \left\Vert\nabla F(\theta^{t})\right\Vert^{2} \leq \frac{1}{ \mathcal{T}}\left(\frac{F(\theta^{0}) - F(\theta^{*})}{\mathcal{A}_{1}} + \mathcal{A}_{2}\sum_{t=0}^{\mathcal{T}-1}\sum_{k=1}^{N}\omega_{k}^{t}\sigma_{k}^{2}\right) + \mathbf{\Phi},
\end{aligned}
\end{equation}
where $\mathcal{A}_{1} = R\eta \left(\frac{1}{2} - 30L^{2}R^{2}\eta^{2} \right)$, $\mathcal{A}_{2} = \frac{60L^{2}R^{3}\eta^{3} + 2R\eta}{R\eta\left(1 - 60L^{2}R^{2}\eta^{2}\right)}$ and $\mathbf{\Phi} = \frac{\left( 10L^{2}R\eta^{2} + L\eta\right)\sigma^{2}}{1 - 60L^{2}R^{2}\eta^{2}}$.

$\hfill \blacksquare$
\subsection{Regularization Terms in the Objective Function}
\label{regularization_appendix}

The proposed method for estimating clients' data heterogeneity relies on the properties of gradient computed for the cross-entropy loss objective. However, the method also applies to the FL algorithms other than FedAvg, in particular those that add a regularization term to combat overfitting, including FedProx \citep{fedprox}, FedDyn\citep{feddyn} and Moon \citep{moon}. In the following discussion, we demonstrate that HiCS-FL remains capable of distinguishing between clients with imbalanced and balanced data when using these other FL algorithms.

\subsubsection{FedProx}
The objective function used by FedProx \citep{fedprox} is
\begin{equation}
\mathcal{L}_{\text{prox}}^{ r} = \mathcal{L}_{\mathbf{CE}}^{r} + \frac{\mu}{2}\left\Vert \theta_{k}^{t,r} - \theta^{t}\right\Vert^{2},
\end{equation}
where $\theta_{k}^{t,r}$ is the vector of client $k$'s local model parameters in the $r$-th local epoch at global round $t$. Therefore, 
contribution of sample $(\mathbf{x}^{(j,n)}, y^{(j,n)})$ to the gradient of $\mathcal{L}_{\text{prox}}$ in local epoch $r$ is
\begin{equation}
\label{grad_prox}
    \frac{\partial \mathcal{L}_{\text{prox}}^{(j,n,r)}}{\partial b_{i}} = \frac{\partial \mathcal{L}_{\textbf{CE}}^{(j,n,r)}}{\partial b_{i}} + \mu\left(b_{i}^{t,r} - b_{i}^{t}\right),
\end{equation}
where $\mathbf{b}^{t,r} = [b_{1}^{t,r},\dots,b_{C}^{t,r}]$ denotes parameters of bias in the output layer of the local model, and $\mathbf{b}^{t} = [b_{1}^{t},\dots,b_{C}^{t}]$ denotes parameters of the global model at round $t$. We assume the model is trained by SGD as the optimizer, and hence
\begin{equation}
    b_{i}^{t,r} - b_{i}^{t} = b_{i}^{t,r-1}  - \eta_{t} \frac{\partial \mathcal{L}_{\text{prox}}^{(j,n,r-1)}}{\partial b_{i}} - b_{i}^{t} =  - \eta_{t} \frac{\partial \mathcal{L}_{\textbf{CE}}^{(j,n,r-1)}}{\partial b_{i}} +  (1 - \eta_{t}\mu)(b_{i}^{t,r-1} - b_{i}^{t}). 
\end{equation}
Therefore,
\begin{equation}
 \begin{aligned}
    b_{i}^{t,r} - b_{i}^{t} &= -\eta_{t}\sum_{s=1}^{r-1} (1 - \eta_{t}\mu)^{r-1 - s} \frac{\partial \mathcal{L}_{\textbf{CE}}^{(j,n,s)}}{\partial b_{i}} + (1 - \eta_{t}\mu)^{r-1}(b_{i}^{t} -b_{i}^{t})\\
    &= -\eta_{t}\sum_{s=1}^{r-1} (1 - \eta_{t}\mu)^{r-1 - s} \frac{\partial \mathcal{L}_{\textbf{CE}}^{(j,n,s)}}{\partial b_{i}},
\end{aligned}   
\end{equation}

and thus
\begin{equation}
     \frac{\partial \mathcal{L}_{\text{prox}}^{(j,n,r)}}{\partial b_{i}} = \frac{\partial \mathcal{L}_{\textbf{CE}}^{(j,n,r)}}{\partial b_{i}}-\eta_{t}\mu\sum_{s=1}^{r-1} (1 - \eta_{t}\mu)^{r-1 - s} \frac{\partial \mathcal{L}_{\textbf{CE}}^{(j,n,s)}}{\partial b_{i}}.
\end{equation}
Taking expectation of both sides yields
\begin{equation}
\begin{aligned}
    \frac{1}{Bl}\sum_{j=1}^{l}\sum_{n=1}^{B}\sum_{r=1}^{R}\mathbb{E}\left[\frac{\partial \mathcal{L}_{\text{prox}}^{(j,n,r)}}{\partial b_{i}}\right] &= \left(-\mathbb{E}[\mathbb{I}\{i = y^{(j,n)}\}]\sum_{c\not = i}\mathcal{E}_{c} + \mathbb{E}[\mathbb{I}(i \not = y^{(j,n)})]\mathcal{E}_{i}\right) \\
    & \cdot \sum_{r=1}^{R}\left(1 -\eta_{t} \mu \sum_{s=1}^{r-1} (1-\eta_{t}\mu)^{r-1-s}\right)\\
    &= \sum_{r=1}^{R}\left(-D_{i}^{(k)}\sum_{c\not = i}\mathcal{E}_{c}  + (1 - D_{i}^{(k)})\mathcal{E}_{i}\right)\left(1 - \eta_{t}\mu\frac{1 - (1 - \eta_{t}\mu)^{r-1}}{\eta_{t}\mu}\right)\\
    &= \sum_{r=1}^{R}c^{r}\left(-D_{i}^{(k)}\sum_{c = 1}^{C}\mathcal{E}_{c}  + \mathcal{E}_{i}\right),
\end{aligned}
\end{equation}
where $c^{r} = (1 - \eta_{t}\mu)^{r-1} > 0$ provided $\eta_{t}$ and $\mu$ are sufficiently small. Therefore, the expectation of the local update of bias in the output layer satisfies 
\begin{equation}
\label{expect_prox}
    \mathbb{E}\left[\Delta b_{i}^{(k)}\right] = C\eta_{t}\left(D_{i}^{(k)}\sum_{c = 1}^{C}\mathcal{E}_{c}  - \mathcal{E}_{i}\right),
\end{equation}
where $C = \sum_{r = 1}^{R} c^{r}$. Eq. (\ref{expect_prox}) is similar to the expression for the expectation of the local updates of bias when applying FedAvg presented in the main paper; clearly, the analysis of HiCS-FL done in the context of FedAvg extends to FedProx.

\subsubsection{FedDyn}
For FedDyn \citep{feddyn}, the objective function in local epoch $r$ 
at global round $t$ is
\begin{equation}
    \mathcal{L}_{\text{dyn}}^{t,r} = \mathcal{L}_{\mathbf{CE}}^{t,r} - \left<\nabla  \mathcal{L}_{\text{dyn}}^{t-1, R},\theta_{k}^{t,r} \right> + \frac{\mu}{2}\left\Vert \theta_{k}^{t,r} - \theta^{t}\right\Vert^{2},
\end{equation}
where $R$ denotes the total number of local epochs. The first order condition for local optima implies
\begin{equation}
    \nabla  \mathcal{L}_{\text{dyn}}^{t, r} - \nabla  \mathcal{L}_{\text{dyn}}^{t-1, R} + \mu(\theta_{k}^{t,r} - \theta^{t}) = 0,
\end{equation}
and, therefore,
\begin{equation}
\begin{aligned}
\label{grad_dyn}
    \frac{\partial \mathcal{L}_{\text{dyn}}^{t,r}}{\partial b_{i}} &= \frac{\partial \mathcal{L}_{\text{dyn}}^{t-1,R}}{\partial b_{i}} - \mu\left(b_{i}^{t,r} - b_{i}^{t}\right)\\
    &= \frac{\partial \mathcal{L}_{\text{dyn}}^{t-2,R}}{\partial b_{i}} - \mu\left(b_{i}^{t-1,R} - b_{i}^{t-1}\right) - \mu\left(b_{i}^{t,r} - b_{i}^{t}\right)\\
    &= -\mu\sum_{\tau=1}^{t-1}\left(b_{i}^{\tau,R} - b_{i}^{\tau}\right) - \mu\left(b_{i}^{t,r} - b_{i}^{t}\right)\\
    &= -\mu\sum_{\tau=1}^{t-1}\Delta b_{i}^{\tau} - \mu\left(b_{i}^{t,r} - b_{i}^{t}\right)\\
    &=  -\mu\sum_{\tau=1}^{t-1}\Delta b_{i}^{\tau} - \mu\left(-\eta_{t}\frac{\partial \mathcal{L}_{\text{dyn}}^{t,r-1}}{\partial b_{i}} + b_{i}^{t,r-1} - b_{i}^{t} \right)\\
    &= -\mu\sum_{\tau=1}^{t-1}\Delta b_{i}^{\tau} + \mu\eta_{t}\left(\sum_{s=1}^{r-1}\frac{\partial \mathcal{L}_{\text{dyn}}^{t,s}}{\partial b_{i}}  \right),
\end{aligned}
\end{equation}
where $\mathbf{b}^{t,r} = [b_{1}^{t,r},\dots,b_{C}^{t,r}]$ denotes the bias parameters in the output layer of the local model at local epoch $r$, and where $\Delta \mathbf{b}^{\tau} = [\Delta b_{1}^{\tau}, \dots, \Delta b_{C}^{\tau}]$ is the local update of the bias at round $\tau$. Since
\begin{equation}
    \frac{\partial \mathcal{L}_{\text{dyn}}^{t,1}}{\partial b_{i}} = -\mu\sum_{\tau=1}^{t-1}\Delta b_{i}^{\tau},
\end{equation}
it holds that 
\begin{equation}
    \frac{\partial \mathcal{L}_{\text{dyn}}^{t,2}}{\partial b_{i}} = -\mu\sum_{\tau=1}^{t-1}\Delta b_{i}^{\tau} + \mu\eta_{t}\left(-\mu\sum_{\tau=1}^{t-1}\Delta b_{i}^{\tau} \right) = -\mu(1 + \mu\eta_{t})\sum_{\tau=1}^{t-1}\Delta b_{i}^{\tau}
\end{equation}
and
\begin{equation}
    \frac{\partial \mathcal{L}_{\text{dyn}}^{t,3}}{\partial b_{i}} = -\mu\sum_{\tau=1}^{t-1}\Delta b_{i}^{\tau} + \mu\eta_{t}\left(-\mu\sum_{\tau=1}^{t-1}\Delta b_{i}^{\tau} - (\mu + \mu^{2}\eta_{t})\sum_{\tau=1}^{t-1}\Delta b_{i}^{\tau}\right) = -\mu(1 + \mu\eta_{t})^{2}\sum_{\tau=1}^{t-1}\Delta b_{i}^{\tau}.
\end{equation}
By induction,
\begin{equation}
    \frac{\partial \mathcal{L}_{\text{dyn}}^{t,r}}{\partial b_{i}} = -\mu(1 + \mu\eta_{t})^{r-1}\sum_{\tau=1}^{t-1}\Delta b_{i}^{\tau}.
\end{equation}
Therefore, the expectation of the local update of bias in the output layer at round $t$ can be computed as
\begin{align}
    \mathbb{E}\left[ \Delta b_{i}^{(k), t}\right] &= \sum_{r=1}^{R}(1 + \mu\eta_{t})^{r-1}\mu\eta_{t}\sum_{\tau=1}^{t-1}\mathbb{E}\left[\Delta b_{i}^{(k),\tau}\right]\\
    &= \left((1+\mu\eta_{t})^{R} - 1 \right)\sum_{\tau=1}^{t-1}\mathbb{E}\left[\Delta b_{i}^{(k),\tau}\right].
\end{align}
Since the objective function of $\mathbb{E}\left[\Delta b_{i}^{(k),1}\right]$ coincides with that of FedAvg,
\begin{equation}
    \mathbb{E}\left[\Delta b_{i}^{(k),1}\right] = \eta_{1}R\left(D_{i}^{(k)}\sum_{c = 1}^{C}\mathcal{E}_{c}  - \mathcal{E}_{i}\right),
\end{equation}
where $\eta_{1}$ is the learning rate at global round $t=1$. Then,
\begin{align}
\mathbb{E}\left[\Delta b_{i}^{(k),2}\right] &= \eta_{1}R\left( (1+\mu\eta_{2})^{R} -1\right)\left(D_{i}^{(k)}\sum_{c = 1}^{C}\mathcal{E}_{c}  -\mathcal{E}_{i}\right)\\
&= a_{1}a_{2}\left(D_{i}^{(k)}\sum_{c = 1}^{C}\mathcal{E}_{c}  -\mathcal{E}_{i}\right),
\end{align}
where $a_{1} = \eta_{1}R$ and $a_{2} =  (1+\mu\eta_{2})^{R} -1$. Furthermore,
\begin{equation}
    \mathbb{E}\left[\Delta b_{i}^{(k),3}\right] = a_1a_{3}( 1+ a_2)\left(D_{i}^{(k)}\sum_{c = 1}^{C}\mathcal{E}_{c}  -\mathcal{E}_{i}\right),
\end{equation}
\begin{equation}
    \mathbb{E}\left[\Delta b_{i}^{(k),4}\right] = a_1a_{4}(1 + a_2 + a_3 + a_2a_3)\left(D_{i}^{(k)}\sum_{c = 1}^{C}\mathcal{E}_{c}  - \mathcal{E}_{i}\right),
\end{equation}
and
\begin{equation}
    \mathbb{E}\left[\Delta b_{i}^{(k),5}\right] = a_1a_{5}(1 + a_2 + a_3 + a_4+ a_2 a_3 + a_3 a_4 + a_2 a_3 a_4)\left(D_{i}^{(k)}\sum_{c = 1}^{C}\mathcal{E}_{c}  - \mathcal{E}_{i}\right).
\end{equation}
By induction,
\begin{align}
\label{eq_dyn}
    \mathbb{E}\left[\Delta b_{i}^{(k),t}\right] &= \left(D_{i}^{(k)}\sum_{c = 1}^{C}\mathcal{E}_{c}  - \mathcal{E}_{i}\right)a_1 a_{t} \cdot 
    \left(1 +  \sum_{i = 0}^{t-3}\sum_{\tau=2}^{t-1}\mathbb{I}(\tau + i < t)\prod_{i=\tau}^{\tau + i } a_{s}\right)\\
    &= a\left(D_{i}^{(k)}\sum_{c = 1}^{C}\mathcal{E}_{c}  -\mathcal{E}_{i}\right),
\end{align}
where $a_{t} = (1+\mu\eta_{t})^{R} - 1$ and $a = a_{1}a_{t}\left(1 +  \sum_{i = 0}^{t-3}\sum_{\tau=2}^{t-1}\mathbb{I}(\tau + i < t)\prod_{i=\tau}^{\tau + i } a_{s}\right) >0$. After comparing Eq. (\ref{eq_dyn}) with its counterpart in the case of FedAvg, we conclude that the previously presented analysis of HiCS-FL extends to FedDyn.

\subsubsection{Model-Contrastive Federated Learning
 (Moon)}

Moon \citep{moon} relies on the objective function with a contrastive term
\begin{equation}    \mathcal{L}_{\text{moon}} = \frac{1}{Bl}\sum_{j=1}^{l}\sum_{n=1}^{B}\mathcal{L}^{(j,n)}_{\mathbf{CE}} - \mu\log \frac{\exp(\text{sim}(\mathbf{z}^{(j,n)},\mathbf{z}_{\text{glob}}^{(j,n)})/T)}{\exp(\text{sim}(\mathbf{z}^{(j,n)},\mathbf{z}_{\text{glob}}^{(j,n)})/T) + \exp(\text{sim}(\mathbf{z}^{(j,n)},\mathbf{z}_{\text{prev}}^{(j,n)})/T)},
\end{equation}
where $\mathbf{z}^{(j,n)}$ denotes the output of the feature extractor of the local model $\theta_{k}^{t}$, $\mathbf{z}_{\text{glob}}^{(j,n)}$ is the output of the feature extractor of the global model $\theta^{t}$, and $\mathbf{z}_{\text{prev}}^{(j,n)}$ is the output of the feature extractor of the local model in the previous round $\theta_{k}^{t-1}$. Since the contrastive term does not depend on the parameters of bias in the output layer, it holds that
\begin{equation}
\label{grad_moon}
    \frac{\partial \mathcal{L}_{\text{moon}}^{(j,n)}}{\partial b_{i}} = \frac{\partial \mathcal{L}_{\textbf{CE}}^{(j,n)}}{\partial b_{i}}. 
\end{equation}
Since the expectation of the local updates of bias in the output layer coincides with the one in case of FedAvg, previously presented analysis of HiCS-FL extends to Moon.

\subsection{Optimization Algorithms Beyond SGD}
\label{optimizers_appendix}

Optimizers beyond SGD utilize different model update rules which in principle may lead to different properties of the local update of the bias in the output layer. However, for several variants of SGD, the properties of the local update of the bias remain such that our presented analysis still applies.
\subsubsection{SGD with momentum}
In each local epoch $r$, SGD with momentum updates the model according to
\begin{equation}
    m_{k}^{t,r} = \mu m_{k}^{t,r-1} +  (1-\mu)\nabla \mathcal{L}_{\textbf{CE}}^{t,r},
\end{equation}
\begin{equation}
    g_{k}^{t,r} = m_{k}^{t,r},
\end{equation}
\begin{equation}
    \theta_{k}^{t,r} = \theta_{k}^{t,r-1} - \eta_{t}g_{k}^{t,r},
\end{equation}
where $ m_{k}^{t,r}$ denotes the momentum in the $r$-th local epoch, $\mu$ is the weight for the momentum, and $m_{k}^{t,1} = \nabla \mathcal{L}_{\textbf{CE}}^{t,1}$. Then
\begin{equation}
    \Delta \theta_{k}^{t} = -\eta_{t}\sum_{r=1}^{R}g_{k}^{t,r},
\end{equation}
where
\begin{equation}
    m_{k}^{t,1} = \nabla \mathcal{L}_{\textbf{CE}}^{t,1},
\end{equation}
\begin{equation}
    m_{k}^{t,2} = \mu \nabla \mathcal{L}_{\textbf{CE}}^{t,1} + (1-\mu)\nabla \mathcal{L}_{\textbf{CE}}^{t,2},
\end{equation}
\begin{equation}
\begin{aligned}
    m_{k}^{t,3} &= \mu \nabla \mathcal{L}_{\textbf{CE}}^{t,2} + (1-\mu)\nabla \mathcal{L}_{\textbf{CE}}^{t,3}\\
    &= \mu^{2} \nabla \mathcal{L}_{\textbf{CE}}^{t,1} + \mu(1-\mu)\nabla \mathcal{L}_{\textbf{CE}}^{t,2} + (1-\mu)\nabla \mathcal{L}_{\textbf{CE}}^{t,3}.
\end{aligned}
\end{equation}

Therefore, 
\begin{equation}
    m_{k}^{t,r} =  \mu^{r-1} \nabla \mathcal{L}_{\textbf{CE}}^{t,1} + (1-\mu)\sum_{\tau = 2}^{r} \mu^{r - \tau} \nabla \mathcal{L}_{\textbf{CE}}^{t,\tau} 
\end{equation}
and thus we have
\begin{equation}
    \Delta \theta_{k}^{t} = -\eta_{t} \left(\sum_{r=2}^{R} \left(\mu^{r-1} \nabla \mathcal{L}_{\textbf{CE}}^{t,1} + (1-\mu)\sum_{\tau = 2}^{r} \mu^{r - \tau} \nabla \mathcal{L}_{\textbf{CE}}^{t,\tau}\right)  + \nabla \mathcal{L}_{\textbf{CE}}^{t,1}\right).
\end{equation}
Similar to the discussion in the previous section,
\begin{align}
    \mathbb{E}\left[ \Delta b_{i}^{(k)}\right] &= \eta_{t}\left(\sum_{r=2}^{R}\left( \mu^{r-1}  + (1-\mu)\sum_{\tau = 2}^{r} \mu^{r - \tau} \right) + 1\right)\left(D_{i}^{(k)}\sum_{c = 1}^{C}\mathcal{E}_{c}  - \mathcal{E}_{i}\right)\\
    &= a\left(D_{i}^{(k)}\sum_{c = 1}^{C}\mathcal{E}_{c}  - \mathcal{E}_{i}\right)
\end{align}
where $a = \eta_{t}\left(\sum_{r=2}^{R}\left( \mu^{r-1}  + (1-\mu)\sum_{\tau = 2}^{r} \mu^{r - \tau} \right) + 1\right) > 0$. Similar result is obtained when SGD applies Nesterov acceleration as long as the optimizers are not using second-order momentum.

\subsubsection{Adam Optimizer}
Recall that the two observations regarding the gradient of $\mathcal{L}_{\textbf{CE}}$ still hold when training the model with an adaptive optimizer such as Adam \citep{adam}. However, Adam updates the model differently from SGD. In particular, each entry of the gradient has an adaptive learning rate tied to its magnitude. With an SGD optimizer, the magnitude of the $i$-th entry of the local update of bias $\Delta \mathbf{b}^{(k)}$ is approximately proportional to the fraction of the samples with label $i$, $D_{i}^{(k)}$ (if $\mathcal{E}_{i}$ is small),
\begin{equation}
    \mathbb{E}\left[\Delta b_{i}^{(k)}\right] = \eta_{t}R\left(D_{i}^{(k)}\sum_{c = 1}^{C}\mathcal{E}_{c}  - \mathcal{E}_{i}\right).
\end{equation}
However, this observation does not hold when using the Adam optimizer for the local update because each entry has a different learning rate $\eta_{t,i}$ and thus
\begin{equation}
    \mathbb{E}\left[\Delta b_{i}^{(k)}\right] = \eta_{t,i}R\left(D_{i}^{(k)}\sum_{c = 1}^{C}\mathcal{E}_{c}  - \mathcal{E}_{i}\right).
\end{equation}
Although the magnitude of $\mathbb{E}\left[\Delta b_{i}^{(k)}\right]$ is no longer approximately proportional to $D_{i}^{(k)}$, we can utilize the sign of $\mathbb{E}\left[\Delta b_{i}^{(k)}\right]$, i.e.,
\begin{equation}
    \text{if } D_{i}^{(k)} \gg D_{j}^{(k)}, \text{then }\mathbb{P}\left( \mathbb{E}\left[\Delta b_{i}^{(k)}\right] > 0\right) \gg \mathbb{P}\left( \mathbb{E}\left[\Delta b_{j}^{(k)}\right] > 0\right).
\end{equation}
Suppose client $k$ has highly imbalanced data, i.e., $H(\mathcal{D}^{(k)})$ is small. Then the maximal component $\max_{i} D_{i}^{(k)}$ is much larger than the other components; in fact, it is likely to have only one positive component in the local update of bias $\Delta \mathbf{b}^{(k)}$. On the contrary, suppose client $u$ has balanced data and thus $H(\mathcal{D}^{(u)})$ is large. The maximal component $\max_{i} D_{i}^{(u)}$ is then very close to the other components, and it is likely to observe larger number of positive components in the local update of $\Delta \mathbf{b}^{(u)}$. While characterizing $\mathbb{P}( \mathbb{E}[\Delta b_{i}^{(k)}] > 0)$ appears challenging, we can empirically infer that client $u$ with more balanced data has a local update of bias $\Delta \mathbf{b}^{(u)}$ with more positive components. With
\begin{equation}
    \hat{H}(\mathcal{D}^{(u)}) \triangleq H(\mbox{softmax}(\Delta\mathbf{b}^{(u)}, T)),
\end{equation}
\begin{equation}
    \hat{H}(\mathcal{D}^{(k)}) \triangleq H(\mbox{softmax}(\Delta\mathbf{b}^{(k)}, T)),
\end{equation}
$\hat{H}(\mathcal{D}^{(u)})$ is more likely to be larger than $\hat{H}(\mathcal{D}^{(k)})$. The examples of estimated entropy when utilizing Adam as the optimizer are provided in Section. \ref{example}.

\subsection{Visualization of Data Partitions}
\label{visualization_appendix}

To generate non-IID data partitions we follow the strategy in \citep{yurochkin2019bayesian}, utilizing Dirichlet distribution with different concentration parameters $\alpha$ to control the heterogeneity
levels. In particular, the number of samples with label $i$ owned by client $k$ is set to $\frac{X_{i}^{(k)}N_{i}}{\sum_{j=1}^{N}X_{i}^{(j)}}$,where $X_{i}^{(1)}, \dots, X_{i}^{(N)}$ are drawn from $\text{Dir}(\alpha)$ and $N_i$ denotes the total number of samples with label $i$ in the overall dataset. For the setting with multiple $\alpha$, we divide the overall training set into $|\alpha|$ equal parts and generate data partitions according to the method above. Figures~\ref{data_cifar10} and \ref{data_mini} illustrate the class distribution of local clients by displaying the number of samples with different labels; colors distinguish between magnitudes -- the darker the color, the more samples are in the class.

\begin{figure*}[h]
    \centering
	  \subfloat[]{
       \includegraphics[width=0.3\linewidth]{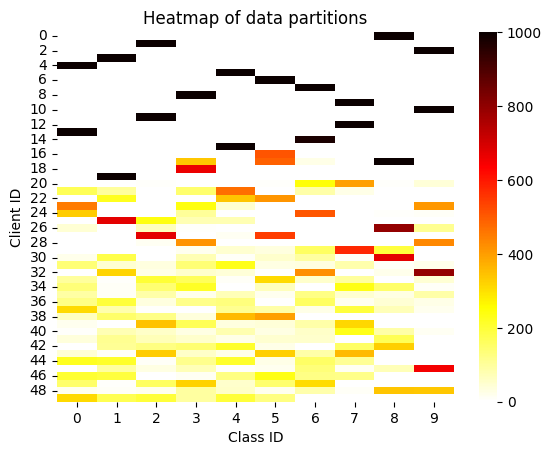}}
	  \subfloat[]{
        \includegraphics[width=0.3\linewidth]{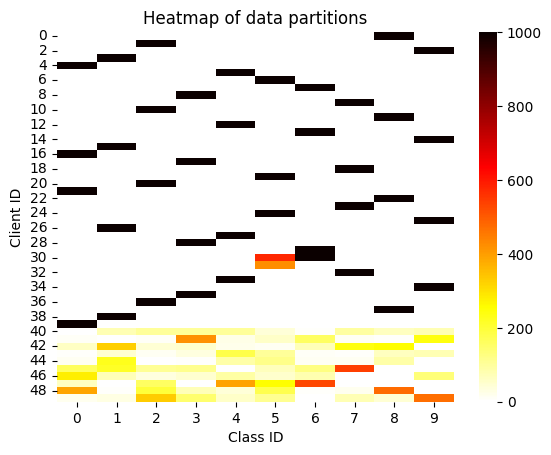}}
	  \subfloat[]{
        \includegraphics[width=0.3\linewidth]{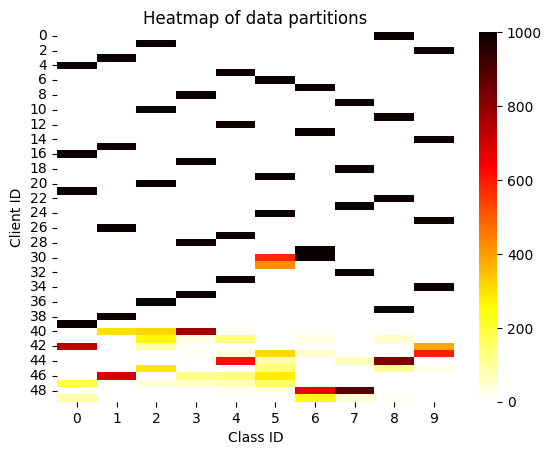}}
      \caption{Results on CIFAR10. Training data is split into 50 partitions according to a Dirichlet distribution (50 clients). The concentration parameter is as follows: (1) $\alpha  \in \{0.001,0.01,0.1,0.5,1.0\}$; (2) $\alpha  \in \{0.001,0.002,0.005,0.01, 0.5\}$; (3) $\alpha  \in \{0.001,0.002,0.005,0.01, 0.1\}$. The figures (a), (b) and (c) correspond to settings (1), (2) and (3), respectively.}
    \label{data_cifar10}
\end{figure*}

\begin{figure*}[h]
    \centering
	  \subfloat[]{
       \includegraphics[width=0.35\linewidth]{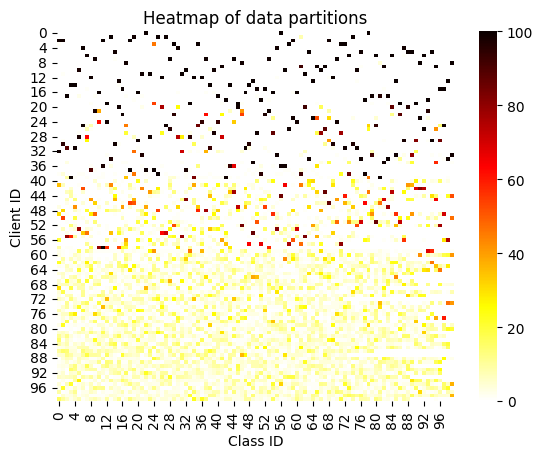}}
	  \subfloat[]{
        \includegraphics[width=0.35\linewidth]{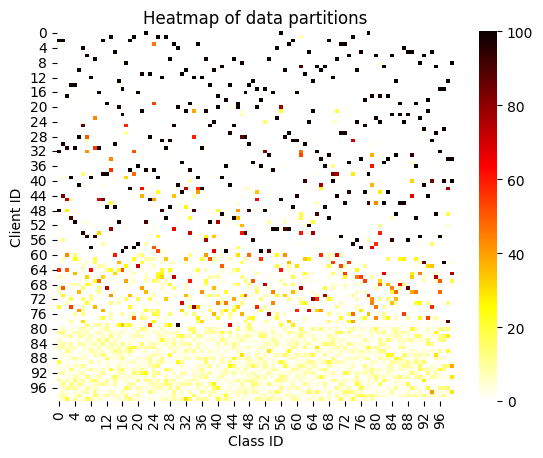}}
	  \caption{Results on Mini-ImageNet. Training data is split into 100 partitions according to Dirichlet distribution (100 clients). The concentration parameter is varied as follows: (1) $\alpha  \in \{0.001,0.01,0.1,0.5, 1.0\}$; (2) $\alpha \in  \{0.001,0.005,0.01,0.1, 1.0\}$. The figures (a) and (b) correspond to settings (1) and (2), respectively.}
    \label{data_mini}
\end{figure*}

\subsection{Computational and Communication Complexity}
\label{complexity}
\begin{table*}[h]
\caption{The columns “Extra Computation” and “Extra Communication” denote the computation and
communication complexity of additional operations in each sampling scheme compared to random sampling. } 
\centering
\begin{tabular}{lcc}
\bottomrule[1pt]
\label{table3}
Method&Extra Computation  &  Extra Communication \\
\hline  
Random     & - &  -   \\
pow-d  & $\mathcal{O}(\left| \theta^{t}\right|)$ &$\mathcal{O}(\left| \theta^{t}\right|)$ \\
CS      & $\mathcal{O}(\left| \theta^{t}\right|)$ &- \\
DivFL   &  $\mathcal{O}(\left| \theta^{t}\right|)$&$\mathcal{O}(\left| \theta^{t}\right|)$ \\
FedCor   & $\mathcal{O}(\left| \theta^{t}\right|)$& - \\
\textbf{HiCS-FL} & $\mathcal{O}(C)$ & -  \\
\toprule[1pt]
\end{tabular}
\end{table*}

We compare the communication and computational costs of
HiCS-FL with those of the competing methods, including Power of Choice (pow-d) \citep{powerofchoice}, Clustered Sampling \citep{clustered} and DivFL \citep{diverse}, and map them against random sampling, as shown in Table.~\ref{table3}. In its ideal setting, pow-d selects $K$ clients with the largest local validation loss among all $N$ clients. To compute the local validation loss at the beginning of a global training round $t$, the server must send the global model to all clients. Compared to the random sampling strategy where the global model is sent to only $K$ clients, pow-d must transmit additional $(N-K) |\theta^{t}|$ model parameters. Moreover, pow-d requires all clients to compute validation loss of the global model $\theta^{t}$ on local datasets, which incurs additional $\mathcal{O}(N |\theta^{t}|)$ computations. While communication requirements of Clustered Sampling do not exceed those of random sampling, the server must run a clustering algorithm on the local updates of dimension $|\theta^{t}|$ (the same as gradients). DivFL relies on maximizing a submodular function to select the most diverse clients based on all clients' gradients, leading to a transmission overhead and additional computation involving $|\theta^{t}|$-dimensional gradients. In our experiments, DivFL has consistently required the longest training time and memory usage due to its dependence on the submodularity maximizer. FedCor \citep{fedcor} cliams that only partial clients participating in the global update after warm-up stage but still needs all clients to perform inference for computing validation loss in the warm-up stage. Our proposed method, HiCS-FL, does not require any additional transmission of model parameters; furthermore, in HiCS-FL the server clusters clients based on their local updates of the bias in the output layer, which is low-dimensional and model-agnostic. Overall, HiCS-FL requires negligible computational overhead to significantly improve the performance of non-iid Federated Learning.

\subsection{Examples of Estimated Entropy}
\label{example}

To further illustrate the proposed framework, here we show a comparison between the estimated entropy of data label distribution and the true entropy. Specifically, Figures \ref{FMNIST} and \ref{mini} show that the entropy estimated by the proposed method is close to the true entropy; the experiments were conducted on FMNIST and Mini-ImageNet, using SGD and Adam as optimizers, respectively. As stated in Theorem \ref{theorem_estimation}, the clients with larger true entropy are likely to have lager estimated entropy. In case where the model is trained with Adam, estimated entropy of data label distribution is not as accurate as in the case of using SGD. Figures \ref{cifar10_1} and \ref{cifar10_2} compare the performance of estimating entropy with SGD and Adam optimizers for the same setting of $\alpha$. Notably, as shown in the figures, the method is capable of distinguishing clients with extremely imbalanced data from those with balanced data.

\begin{figure*}[h]
    \centering
	  \subfloat[]{
       \includegraphics[width=0.3\linewidth]{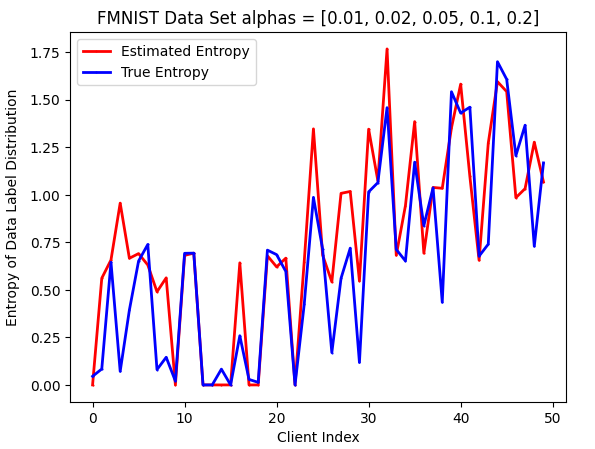}}
       \hspace{0.3 in}
	  \subfloat[]{
        \includegraphics[width=0.28\linewidth]{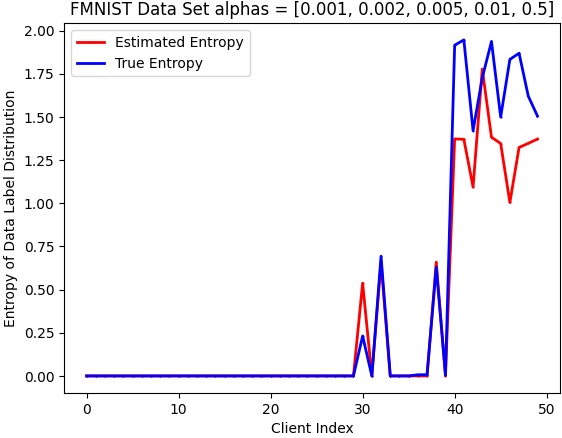}}
	  \caption{The estimated entropy of data label distribution in experiments on FMNIST with SGD as the optimizer. The parameter $\alpha$ for the two figures: (a) $\alpha \in \{0.01, 0.02, 0.05, 0.1, 0.2\}$; (b) $\alpha \in \{0.001, 0.002, 0.005, 0.01, 0.5\}$}
    \label{FMNIST}
\end{figure*}
\begin{figure*}[!h]
    \centering
        \subfloat[]{
        \includegraphics[width=0.3\linewidth]{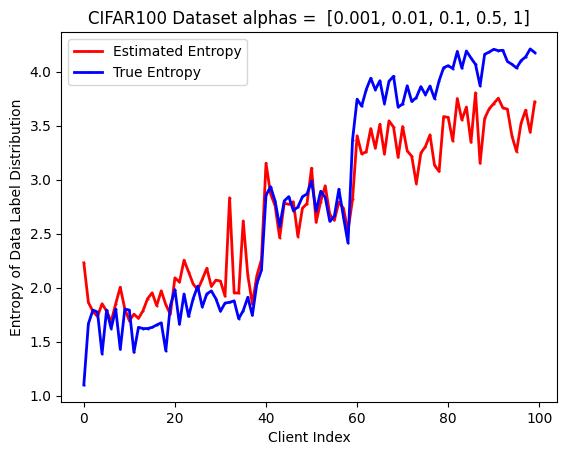}}
        \hspace{0.3 in}
	  \subfloat[]{
       \includegraphics[width=0.3\linewidth]{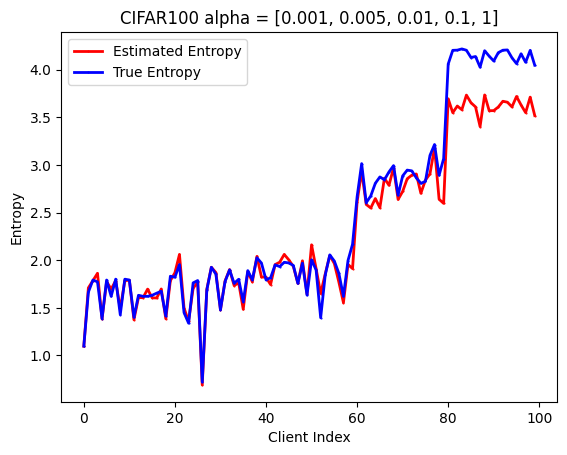}}
	  
	  \caption{The estimated entropy of data label distribution in experiments on Mini-ImageNet with Adam as the optimizer. The parameter $\alpha$ for the two figures: (a) $\alpha \in \{0.001, 0.01, 0.1, 0.5, 1.0\}$; (b) $\alpha \in \{0.001, 0.005, 0.01, 0.1, 1.0\}$.}
    \label{mini}
\end{figure*}
\begin{figure}[H]
    \centering
	  \subfloat[]{
       \includegraphics[width=0.3\linewidth]{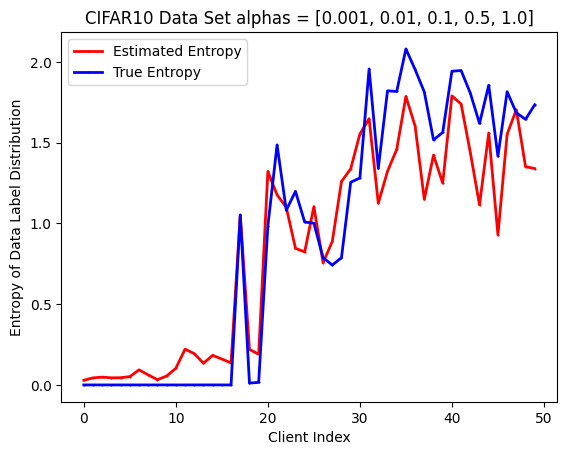}}
       \hspace{0.3 in}
	  \subfloat[]{
        \includegraphics[width=0.3\linewidth]{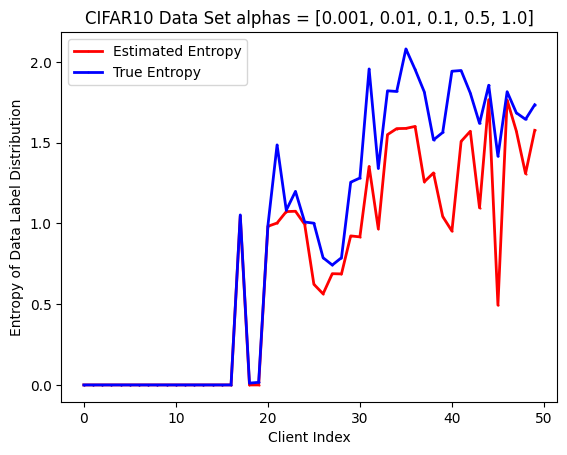}}
	  \caption{The estimated entropy of data label distribution in experiments on CIFAR10 with $\alpha \in \{0.001, 0.01, 0.1, 0.5, 1.0\}$. (a) The result of the experiments using SGD as the optimizer. (b) The result of the experiments using Adam as the optimizer.}
    \label{cifar10_1}
\end{figure}
\begin{figure}[H]
    \centering
	  \subfloat[]{
       \includegraphics[width=0.3\linewidth]{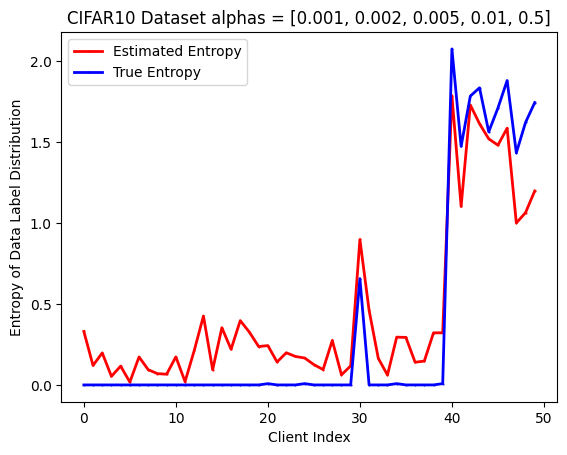}}
       \hspace{0.3 in}
	  \subfloat[]{
        \includegraphics[width=0.3\linewidth]{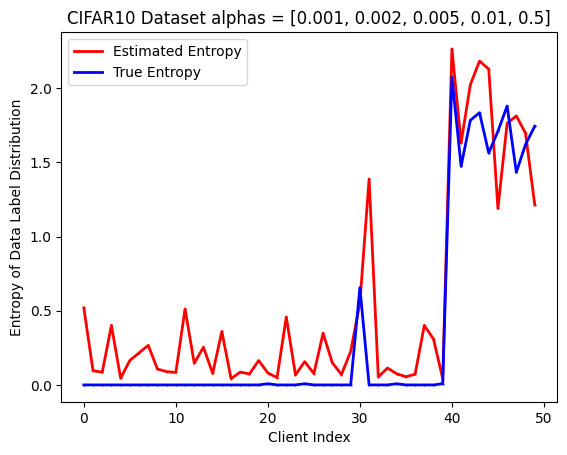}}
	  \caption{The estimated entropy of data label distribution in experiments on CIFAR10 with $\alpha \in $ \{$0.001$, $0.002$, $0.00$5, $0.01$, $0.5$\}. (a) The result of the experiments using SGD as the optimizer. (b) The result of the experiments using Adam as the optimizer.}
    \label{cifar10_2}
\end{figure}


\newpage
\section*{NeurIPS Paper Checklist}
\begin{enumerate}

\item {\bf Claims}
    \item[] Question: Do the main claims made in the abstract and introduction accurately reflect the paper's contributions and scope?
    \item[] Answer: \answerYes{} 
    \item[] Justification: \answerNA{} 
    \item[] Guidelines:
    \begin{itemize}
        \item The answer NA means that the abstract and introduction do not include the claims made in the paper.
        \item The abstract and/or introduction should clearly state the claims made, including the contributions made in the paper and important assumptions and limitations. A No or NA answer to this question will not be perceived well by the reviewers. 
        \item The claims made should match theoretical and experimental results, and reflect how much the results can be expected to generalize to other settings. 
        \item It is fine to include aspirational goals as motivation as long as it is clear that these goals are not attained by the paper. 
    \end{itemize}

\item {\bf Limitations}
    \item[] Question: Does the paper discuss the limitations of the work performed by the authors?
    \item[] Answer: \answerYes{} 
    \item[] Justification: \answerNA{}
    \item[] Guidelines:
    \begin{itemize}
        \item The answer NA means that the paper has no limitation while the answer No means that the paper has limitations, but those are not discussed in the paper. 
        \item The authors are encouraged to create a separate "Limitations" section in their paper.
        \item The paper should point out any strong assumptions and how robust the results are to violations of these assumptions (e.g., independence assumptions, noiseless settings, model well-specification, asymptotic approximations only holding locally). The authors should reflect on how these assumptions might be violated in practice and what the implications would be.
        \item The authors should reflect on the scope of the claims made, e.g., if the approach was only tested on a few datasets or with a few runs. In general, empirical results often depend on implicit assumptions, which should be articulated.
        \item The authors should reflect on the factors that influence the performance of the approach. For example, a facial recognition algorithm may perform poorly when image resolution is low or images are taken in low lighting. Or a speech-to-text system might not be used reliably to provide closed captions for online lectures because it fails to handle technical jargon.
        \item The authors should discuss the computational efficiency of the proposed algorithms and how they scale with dataset size.
        \item If applicable, the authors should discuss possible limitations of their approach to address problems of privacy and fairness.
        \item While the authors might fear that complete honesty about limitations might be used by reviewers as grounds for rejection, a worse outcome might be that reviewers discover limitations that aren't acknowledged in the paper. The authors should use their best judgment and recognize that individual actions in favor of transparency play an important role in developing norms that preserve the integrity of the community. Reviewers will be specifically instructed to not penalize honesty concerning limitations.
    \end{itemize}

\item {\bf Theory Assumptions and Proofs}
    \item[] Question: For each theoretical result, does the paper provide the full set of assumptions and a complete (and correct) proof?
    \item[] Answer: \answerYes{} 
    \item[] Justification: \answerNA{}
    \item[] Guidelines:
    \begin{itemize}
        \item The answer NA means that the paper does not include theoretical results. 
        \item All the theorems, formulas, and proofs in the paper should be numbered and cross-referenced.
        \item All assumptions should be clearly stated or referenced in the statement of any theorems.
        \item The proofs can either appear in the main paper or the supplemental material, but if they appear in the supplemental material, the authors are encouraged to provide a short proof sketch to provide intuition. 
        \item Inversely, any informal proof provided in the core of the paper should be complemented by formal proofs provided in appendix or supplemental material.
        \item Theorems and Lemmas that the proof relies upon should be properly referenced. 
    \end{itemize}

    \item {\bf Experimental Result Reproducibility}
    \item[] Question: Does the paper fully disclose all the information needed to reproduce the main experimental results of the paper to the extent that it affects the main claims and/or conclusions of the paper (regardless of whether the code and data are provided or not)?
    \item[] Answer: \answerYes{} 
    \item[] Justification: \answerNA{}
    \item[] Guidelines:
    \begin{itemize}
        \item The answer NA means that the paper does not include experiments.
        \item If the paper includes experiments, a No answer to this question will not be perceived well by the reviewers: Making the paper reproducible is important, regardless of whether the code and data are provided or not.
        \item If the contribution is a dataset and/or model, the authors should describe the steps taken to make their results reproducible or verifiable. 
        \item Depending on the contribution, reproducibility can be accomplished in various ways. For example, if the contribution is a novel architecture, describing the architecture fully might suffice, or if the contribution is a specific model and empirical evaluation, it may be necessary to either make it possible for others to replicate the model with the same dataset, or provide access to the model. In general. releasing code and data is often one good way to accomplish this, but reproducibility can also be provided via detailed instructions for how to replicate the results, access to a hosted model (e.g., in the case of a large language model), releasing of a model checkpoint, or other means that are appropriate to the research performed.
        \item While NeurIPS does not require releasing code, the conference does require all submissions to provide some reasonable avenue for reproducibility, which may depend on the nature of the contribution. For example
        \begin{enumerate}
            \item If the contribution is primarily a new algorithm, the paper should make it clear how to reproduce that algorithm.
            \item If the contribution is primarily a new model architecture, the paper should describe the architecture clearly and fully.
            \item If the contribution is a new model (e.g., a large language model), then there should either be a way to access this model for reproducing the results or a way to reproduce the model (e.g., with an open-source dataset or instructions for how to construct the dataset).
            \item We recognize that reproducibility may be tricky in some cases, in which case authors are welcome to describe the particular way they provide for reproducibility. In the case of closed-source models, it may be that access to the model is limited in some way (e.g., to registered users), but it should be possible for other researchers to have some path to reproducing or verifying the results.
        \end{enumerate}
    \end{itemize}

\item {\bf Open access to data and code}
    \item[] Question: Does the paper provide open access to the data and code, with sufficient instructions to faithfully reproduce the main experimental results, as described in supplemental material?
    \item[] Answer: \answerYes{} 
    \item[] Justification: \answerNA{}
    \item[] Guidelines:
    \begin{itemize}
        \item The answer NA means that paper does not include experiments requiring code.
        \item Please see the NeurIPS code and data submission guidelines (\url{https://nips.cc/public/guides/CodeSubmissionPolicy}) for more details.
        \item While we encourage the release of code and data, we understand that this might not be possible, so “No” is an acceptable answer. Papers cannot be rejected simply for not including code, unless this is central to the contribution (e.g., for a new open-source benchmark).
        \item The instructions should contain the exact command and environment needed to run to reproduce the results. See the NeurIPS code and data submission guidelines (\url{https://nips.cc/public/guides/CodeSubmissionPolicy}) for more details.
        \item The authors should provide instructions on data access and preparation, including how to access the raw data, preprocessed data, intermediate data, and generated data, etc.
        \item The authors should provide scripts to reproduce all experimental results for the new proposed method and baselines. If only a subset of experiments are reproducible, they should state which ones are omitted from the script and why.
        \item At submission time, to preserve anonymity, the authors should release anonymized versions (if applicable).
        \item Providing as much information as possible in supplemental material (appended to the paper) is recommended, but including URLs to data and code is permitted.
    \end{itemize}

\item {\bf Experimental Setting/Details}
    \item[] Question: Does the paper specify all the training and test details (e.g., data splits, hyperparameters, how they were chosen, type of optimizer, etc.) necessary to understand the results?
    \item[] Answer: \answerYes{} 
    \item[] Justification: \answerNA{}
    \item[] Guidelines:
    \begin{itemize}
        \item The answer NA means that the paper does not include experiments.
        \item The experimental setting should be presented in the core of the paper to a level of detail that is necessary to appreciate the results and make sense of them.
        \item The full details can be provided either with the code, in appendix, or as supplemental material.
    \end{itemize}

\item {\bf Experiment Statistical Significance}
    \item[] Question: Does the paper report error bars suitably and correctly defined or other appropriate information about the statistical significance of the experiments?
    \item[] Answer: \answerYes{} 
    \item[] Justification: \answerNA{}
    \item[] Guidelines:
    \begin{itemize}
        \item The answer NA means that the paper does not include experiments.
        \item The authors should answer "Yes" if the results are accompanied by error bars, confidence intervals, or statistical significance tests, at least for the experiments that support the main claims of the paper.
        \item The factors of variability that the error bars are capturing should be clearly stated (for example, train/test split, initialization, random drawing of some parameter, or overall run with given experimental conditions).
        \item The method for calculating the error bars should be explained (closed form formula, call to a library function, bootstrap, etc.)
        \item The assumptions made should be given (e.g., Normally distributed errors).
        \item It should be clear whether the error bar is the standard deviation or the standard error of the mean.
        \item It is OK to report 1-sigma error bars, but one should state it. The authors should preferably report a 2-sigma error bar than state that they have a 96\% CI, if the hypothesis of Normality of errors is not verified.
        \item For asymmetric distributions, the authors should be careful not to show in tables or figures symmetric error bars that would yield results that are out of range (e.g. negative error rates).
        \item If error bars are reported in tables or plots, The authors should explain in the text how they were calculated and reference the corresponding figures or tables in the text.
    \end{itemize}

\item {\bf Experiments Compute Resources}
    \item[] Question: For each experiment, does the paper provide sufficient information on the computer resources (type of compute workers, memory, time of exNoecution) needed to reproduce the experiments?
    \item[] Answer: \answerNo{} 
    \item[] Justification: \answerNA{}
    \item[] Guidelines:
    \begin{itemize}
        \item The answer NA means that the paper does not include experiments.
        \item The paper should indicate the type of compute workers CPU or GPU, internal cluster, or cloud provider, including relevant memory and storage.
        \item The paper should provide the amount of compute required for each of the individual experimental runs as well as estimate the total compute. 
        \item The paper should disclose whether the full research project required more compute than the experiments reported in the paper (e.g., preliminary or failed experiments that didn't make it into the paper). 
    \end{itemize}
    
\item {\bf Code Of Ethics}
    \item[] Question: Does the research conducted in the paper conform, in every respect, with the NeurIPS Code of Ethics \url{https://neurips.cc/public/EthicsGuidelines}?
    \item[] Answer: \answerYes{} 
    \item[] Justification: \answerNA{}
    \item[] Guidelines:
    \begin{itemize}
        \item The answer NA means that the authors have not reviewed the NeurIPS Code of Ethics.
        \item If the authors answer No, they should explain the special circumstances that require a deviation from the Code of Ethics.
        \item The authors should make sure to preserve anonymity (e.g., if there is a special consideration due to laws or regulations in their jurisdiction).
    \end{itemize}

\item {\bf Broader Impacts}
    \item[] Question: Does the paper discuss both potential positive societal impacts and negative societal impacts of the work performed?
    \item[] Answer: \answerNA{} 
    \item[] Justification: \answerNA{}
    \item[] Guidelines:
    \begin{itemize}
        \item The answer NA means that there is no societal impact of the work performed.
        \item If the authors answer NA or No, they should explain why their work has no societal impact or why the paper does not address societal impact.
        \item Examples of negative societal impacts include potential malicious or unintended uses (e.g., disinformation, generating fake profiles, surveillance), fairness considerations (e.g., deployment of technologies that could make decisions that unfairly impact specific groups), privacy considerations, and security considerations.
        \item The conference expects that many papers will be foundational research and not tied to particular applications, let alone deployments. However, if there is a direct path to any negative applications, the authors should point it out. For example, it is legitimate to point out that an improvement in the quality of generative models could be used to generate deepfakes for disinformation. On the other hand, it is not needed to point out that a generic algorithm for optimizing neural networks could enable people to train models that generate Deepfakes faster.
        \item The authors should consider possible harms that could arise when the technology is being used as intended and functioning correctly, harms that could arise when the technology is being used as intended but gives incorrect results, and harms following from (intentional or unintentional) misuse of the technology.
        \item If there are negative societal impacts, the authors could also discuss possible mitigation strategies (e.g., gated release of models, providing defenses in addition to attacks, mechanisms for monitoring misuse, mechanisms to monitor how a system learns from feedback over time, improving the efficiency and accessibility of ML).
    \end{itemize}
    
\item {\bf Safeguards}
    \item[] Question: Does the paper describe safeguards that have been put in place for responsible release of data or models that have a high risk for misuse (e.g., pretrained language models, image generators, or scraped datasets)?
    \item[] Answer: \answerNA{} 
    \item[] Justification: \answerNA{}
    \item[] Guidelines:
    \begin{itemize}
        \item The answer NA means that the paper poses no such risks.
        \item Released models that have a high risk for misuse or dual-use should be released with necessary safeguards to allow for controlled use of the model, for example by requiring that users adhere to usage guidelines or restrictions to access the model or implementing safety filters. 
        \item Datasets that have been scraped from the Internet could pose safety risks. The authors should describe how they avoided releasing unsafe images.
        \item We recognize that providing effective safeguards is challenging, and many papers do not require this, but we encourage authors to take this into account and make a best faith effort.
    \end{itemize}

\item {\bf Licenses for existing assets}
    \item[] Question: Are the creators or original owners of assets (e.g., code, data, models), used in the paper, properly credited and are the license and terms of use explicitly mentioned and properly respected?
    \item[] Answer: \answerYes{} 
    \item[] Justification: \answerNA{}
    \item[] Guidelines:
    \begin{itemize}
        \item The answer NA means that the paper does not use existing assets.
        \item The authors should cite the original paper that produced the code package or dataset.
        \item The authors should state which version of the asset is used and, if possible, include a URL.
        \item The name of the license (e.g., CC-BY 4.0) should be included for each asset.
        \item For scraped data from a particular source (e.g., website), the copyright and terms of service of that source should be provided.
        \item If assets are released, the license, copyright information, and terms of use in the package should be provided. For popular datasets, \url{paperswithcode.com/datasets} has curated licenses for some datasets. Their licensing guide can help determine the license of a dataset.
        \item For existing datasets that are re-packaged, both the original license and the license of the derived asset (if it has changed) should be provided.
        \item If this information is not available online, the authors are encouraged to reach out to the asset's creators.
    \end{itemize}

\item {\bf New Assets}
    \item[] Question: Are new assets introduced in the paper well documented and is the documentation provided alongside the assets?
    \item[] Answer: \answerNA{} 
    \item[] Justification: \answerNA{}
    \item[] Guidelines:
    \begin{itemize}
        \item The answer NA means that the paper does not release new assets.
        \item Researchers should communicate the details of the dataset/code/model as part of their submissions via structured templates. This includes details about training, license, limitations, etc. 
        \item The paper should discuss whether and how consent was obtained from people whose asset is used.
        \item At submission time, remember to anonymize your assets (if applicable). You can either create an anonymized URL or include an anonymized zip file.
    \end{itemize}

\item {\bf Crowdsourcing and Research with Human Subjects}
    \item[] Question: For crowdsourcing experiments and research with human subjects, does the paper include the full text of instructions given to participants and screenshots, if applicable, as well as details about compensation (if any)? 
    \item[] Answer: \answerNA{} 
    \item[] Justification: \answerNA{}
    \item[] Guidelines:
    \begin{itemize}
        \item The answer NA means that the paper does not involve crowdsourcing nor research with human subjects.
        \item Including this information in the supplemental material is fine, but if the main contribution of the paper involves human subjects, then as much detail as possible should be included in the main paper. 
        \item According to the NeurIPS Code of Ethics, workers involved in data collection, curation, or other labor should be paid at least the minimum wage in the country of the data collector. 
    \end{itemize}

\item {\bf Institutional Review Board (IRB) Approvals or Equivalent for Research with Human Subjects}
    \item[] Question: Does the paper describe potential risks incurred by study participants, whether such risks were disclosed to the subjects, and whether Institutional Review Board (IRB) approvals (or an equivalent approval/review based on the requirements of your country or institution) were obtained?
    \item[] Answer: \answerNA{} 
    \item[] Justification: \answerNA{}
    \item[] Guidelines:
    \begin{itemize}
        \item The answer NA means that the paper does not involve crowdsourcing nor research with human subjects.
        \item Depending on the country in which research is conducted, IRB approval (or equivalent) may be required for any human subjects research. If you obtained IRB approval, you should clearly state this in the paper. 
        \item We recognize that the procedures for this may vary significantly between institutions and locations, and we expect authors to adhere to the NeurIPS Code of Ethics and the guidelines for their institution. 
        \item For initial submissions, do not include any information that would break anonymity (if applicable), such as the institution conducting the review.
    \end{itemize}

\end{enumerate}

\end{document}